\documentclass[unnumsec,webpdf,modern,large]{mam-authoring-template}%

\graphicspath{{Fig/}}
\usepackage{amsmath,amsfonts,amssymb,amsthm}
\usepackage{mathtools}
\DeclarePairedDelimiterX{\abs}[1]{\lvert}{\rvert}{#1}
\DeclarePairedDelimiterX{\norm}[1]{\lVert}{\rVert}{#1}

\def\q{\mathbf{q}}
\def\rv{\mathbf{r}}

\begin{document}

\copyrightyear{2025}



\title[]{Contrast transfer functions help quantify neural network out-of-distribution generalization in HRTEM}

\author[1,2,$\ast$]{Luis Rangel DaCosta\ORCID{0000-0001-8966-0408}}
\author[1,2, $\ast$]{Mary C. Scott\ORCID{0000-0002-9543-6725}}

\authormark{Luis Rangel DaCosta and Mary C. Scott}

\address[1]{\orgdiv{Department of Materials Science and Engineering}, \orgname{University of California, Berkeley}, \orgaddress{\street{Hearst Memorial Mining Building}, \postcode{94720}, \state{CA}, \country{USA}}}
\address[2]{\orgdiv{National Center for Electron Microscopy, Molecular Foundry}, \orgname{Lawrence Berkeley National Laboratory}, \orgaddress{\street{1 Cyclotron Rd.}, \postcode{94720}, \state{CA}, \country{USA}}}

\corresp[$\ast$]{Corresponding author. \href{email:lrangeldacosta@lbl.gov}{lrangeldacosta@lbl.gov}, \href{email:mary.scott@berkeley.edu}{mary.scott@berkeley.edu}}


\abstract{Neural networks, while effective for tackling many challenging scientific tasks, are not known to perform well out-of-distribution (OOD), i.e., within domains which differ from their training data. Understanding neural network OOD generalization is paramount to their successful deployment in experimental workflows, especially when ground-truth knowledge about the experiment is hard to establish or experimental conditions significantly vary. With inherent access to ground-truth information and fine-grained control of underlying distributions, simulation-based data curation facilitates precise investigation of OOD generalization behavior. Here, we probe generalization with respect to imaging conditions of neural network segmentation models for high-resolution transmission electron microscopy (HRTEM) imaging of nanoparticles, training and measuring the OOD generalization of over 12,000 neural networks using synthetic data generated via random structure sampling and multislice simulation. Using the HRTEM contrast transfer function, we further develop a framework to compare information content of HRTEM datasets and quantify OOD domain shifts. We demonstrate that neural network segmentation models enjoy significant performance stability, but will smoothly and predictably worsen as imaging conditions shift from the training distribution. Lastly, we consider limitations of our approach in explaining other OOD shifts, such as of the atomic structures, and discuss complementary techniques for understanding generalization in such settings.}

\maketitle

\section{Introduction}
The out-of-distribution generalization capability of neural networks underpins the utility of machine learning models in experimental settings. Out-of-distribution (OOD) generalization considers performance on data that differs from the training distribution and we can broadly consider machine learning models to generalize well OOD when they can be used throughout large data domains without meaningful performance degradation. Many key advances in machine learning have worked towards developing an understanding of in-distribution generalization---the ability of a neural network model to learn near-optimal functions for analyzing data similar to their training distribution---but our comparative understanding of OOD generalization lags severely behind \citep{liuOutOfDistributionGeneralizationSurvey2023, yuSurveyEvaluationOutofDistribution2024}, largely relying on empirical evidence and benchmark studies in restricted domains \citep{kangDeepNeuralNetworks2024, yeOoDBenchQuantifyingUnderstanding2022}. Few tools currently exist which can predict and quantitatively describe under which scenarios and to what extent a trained neural network model will successfully generalize to new target domains. As neural network models become increasingly adopted within experimental microscopy workflows, it is becoming increasingly important to devote effort to understanding OOD generalization phenomena as it applies to characterization.

OOD generalization behavior can be explicitly evaluated in supervised learning settings where supervision labels can be used to directly compute model performance, given that suitable OOD benchmark data are accessible. Such benchmark data, however, are not readily available in many experimental settings. Experimental acquisition of domain-shifted benchmark data in transmission electron microscopy (TEM) is feasible along some more controllable aspects of our datasets, like imaging resolution or nanoparticle chemistry \citep{sytwuGeneralizationExperimentalParameters2024}, but is expensive to obtain and annotate. Acquiring and annotating benchmark data for more complex analysis tasks which might be challenging for expert practitioners, or for which aspects of domain shift could be difficult to describe, measure, or calibrate, quickly becomes infeasible. Alternatively, simulation-based data generation approaches \citep{rangeldacostaRobustSyntheticData2024} provide potentially unlimited data with ground-truth annotations and offer an attractive alternative for studying machine learning model generalization. Electron microscopists have recently rapidly adopted neural networks for atomic resolution analysis workflows with the TEM \citep{madsenDeepLearningApproach2018a, groschnerHighThroughputPipeline2020, sadreDeepLearningSegmentation2021a, vincentDevelopingEvaluatingDeep2021, wangAutoDetectmNPUnsupervisedMachine2021, edeDeepLearningElectron2021, munshiDisentanglingMultipleScattering2022a, linTEMImageNetTrainingLibrary2021a, leePtyRADHighperformanceFlexible2025, mccrayAcceleratingIterativePtychography2025, abebeSAMIAmSemanticBoosting2025, kimSelfsupervisedMachineLearning2025, lobatoDeepConvolutionalNeural2024, belardiImprovingMultisliceElectron2025, khanLeveragingGenerativeAdversarial2023a}. Alongside well-established tools for image simulation \citep{abtem, rangeldacostaPrismatic20Simulation2021}, there exists a unique opportunity in electron microscopy to more fully develop our understanding of the limitations of neural network models for experimental data analysis and their potential to generalize across experimental scenarios.

Information theory underlies many of our working models of datasets and learning theory in machine learning and forms the classical basis for understanding OOD generalization of neural network models as a phenomenon arising from domain shift \citep{ben-davidTheoryLearningDifferent2010}. Recent approaches extending such information theoretic perspective with ideas from causal modeling---specifically, the invariance of learnable features \citep{arjovskyInvariantRiskMinimization2020, ahujaInvariancePrincipleMeets2022, yeTheoreticalFrameworkOutofdistribution2021}---have shown promise in providing a unified perspective of OOD generalization across learning tasks. While there exists some purely model- or optimization-centric methods to detect OOD data or improve robustness to OOD shifts \citep{rameDiverseWeightAveraging2023, rameFishrInvariantGradient2022}, such approaches do not readily predict the extent of failure of a model OOD and  often rely on expensive computations. In contrast, data-centric approaches utilizing probabilistic and information theoretic perspectives \citep{cohen-wangAskYourDistribution2024, liuEmpiricalStudyDistribution2022, millerAccuracyLineStrong2021, shiLCAontheLineBenchmarkingOutofDistribution2024} develop an understanding of OOD performance via inter-dataset relationships and their relation to model behavior, laying out a route towards a more direct, mechanistic understanding of the OOD behavior of neural networks used for TEM analysis. 

Within electron microscopy, concepts in information theory most commonly arise when considering information---or contrast---transfer functions. Contrast transfer functions describe how information propagates from samples to measurements via the system of imaging lenses, forming the foundation of our theories of image formation and techniques for interpreting the information contained within micrographs \citep{roseInformationTransferTransmission1984, vandyckUltimateResolutionInformation1992, dejongUltimateResolutionInformation1993, nellistResolutionInformationLimit1995, spenceHighResolutionElectronMicroscopy2017}. Due to the confluence of consumer-available high-performance computing, ultra-fast detectors, and advanced ptychographic reconstruction algorithms \citep{chenElectronPtychographyAchieves2021, kucukogluLowdoseCryoelectronPtychography2024, varnavidesIterativePhaseRetrieval2024} in recent years, understanding information transfer has been considered important for determining limits of the modern TEM \citep{varnavidesIterativePhaseRetrieval2024, varnavidesContrastTransferSpectral2025, maInformation4DSTEMWhere2025}. For TEM imaging data, the contrast transfer function provides a natural setting for understanding image datasets and their information content from a distributional perspective. 

In this work, we take a detailed look at the out-of-distribution generalization behavior of neural network models trained via supervised learning for high-resolution transmission electron microscopy (HRTEM) analysis, focusing on semantic segmentation of atomic-resolution micrographs of nanoparticles on substrates. We begin by seeking to understand neural network OOD behavior with respects to shifts in experimental imaging conditions, which are experimental factors we can describe analytically and control moderately well, but which are hard to precisely measure and can shift significantly between different experiments or TEM operators. Using fully simulated datasets, whose imaging conditions we can arbitrarily control, we demonstrate that atomic resolution segmentation networks have stable and predictable degradation of performance when evaluating OOD data. Moreover, the extent of performance degradation can be mechanistically understood from an information theoretic perspective by relating HRTEM training and test datasets via their respective contrast transfer functions. We further  establish data-centric settings with this comparative framework under which model performance will appreciably degrade. Lastly, we discuss extensions and limitations of our approach for understanding the effect of distribution shifts across different experimental settings, especially when considering shifts in the distribution of atomic structures. 

\section{Methods}

\subsection{Information transfer metrics}

\begin{figure*}[hbtp]
    \centering
    \includegraphics[width=4.8in]{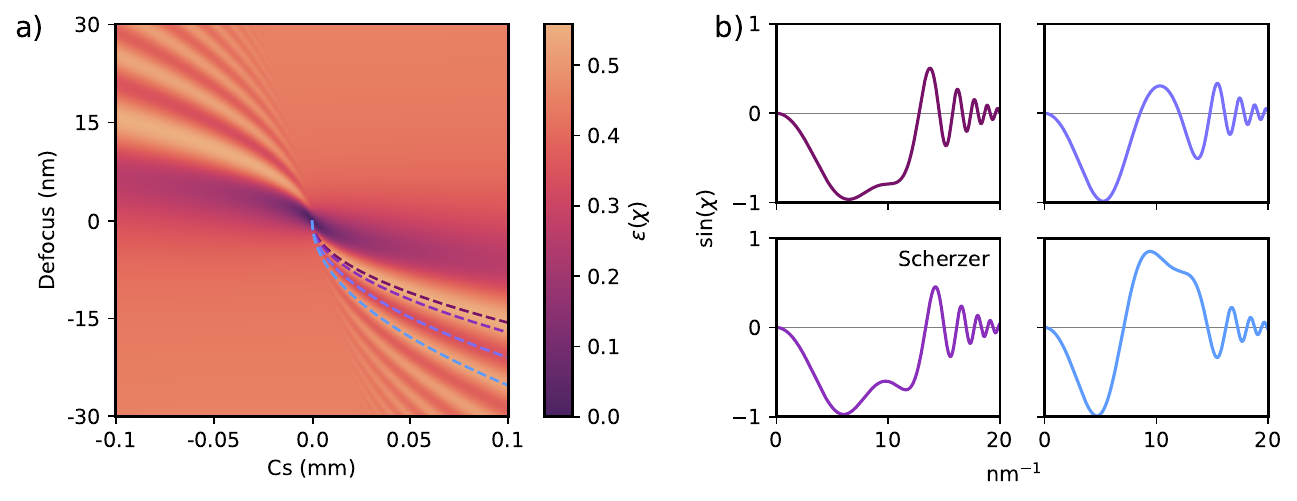}
    \caption{(a) Example visualization of $\epsilon(\chi)$ for aberrations consisting solely of defocus and spherical aberrations, with magnitudes up to 30nm and 0.1mm, respectively, at 300kV and with a chromatic aberration envelope corresponding to a focal spread of 10\r{A}. Bright bands in regions where defocus and spherical aberration have opposite sign correspond to the formation of passbands in the contrast transfer function. (b) A selection of phase contrast transfer functions corresponding to maxima and minima in (a) are visualized for a lens with a spherical aberration coefficient of 25$\mu$m.}
    \label{fig:emetric}
\end{figure*}

We will look to compare the imaging conditions between two HRTEM datasets via their corresponding contrast transfer functions. To simplify the form of the contrast transfer functions, we approximate our sample as a phase object. Throughout, we will use $\rv$ to represent real space coordinates and $\q$ to represent reciprocal space coordinates. 

After propagation through the sample, the exit wavefunction $\psi_f(\q)$ for HRTEM imaging of phase objects can be calculated as 

\begin{equation}
    \psi_f(\q) = \underset{\rv \rightarrow \q}{\mathcal{F}}\left[\psi_0(\rv) \exp(i \sigma V(\rv))\right] H(\q)
\end{equation}

\noindent where $\psi_0(\rv)$ is an incident plane wave, $H(\q)$ is the transfer function, $V(\rv)$ is the electrostatic potential of the sample, and $\underset{\rv \rightarrow \q}{\mathcal{F}}$ is the Fourier transform from real to reciprocal space. $H(\q)$ directly modulates the wavefunction and is the product of several components \citep{williamsTransmissionElectronMicroscopy2008}:

\begin{equation}
    H(\q) = A(\q) E(\q) \exp(-i \chi(\q)).
\end{equation}

\noindent $A(\q)$ is an aperture function, which defines the maximum angular cutoff of incident electrons in the microscope due to a limiting aperture; $E(\q)$ is a damping envelope function, which encompasses high-frequency damping arising from a variety of experimental sources; and $\exp(i\chi(\q))$ is know as the phase contrast transfer function. Since we typically work with measured image intensities $I(\rv) = \psi_f(\rv) \overline{\psi_f(\rv)}$, we instead utilize the intensity transfer function $T(\q)$, which can be be written as 

\begin{equation}
    T(\q) = 2 E(\q) A(\q) \sin(\chi(\q)).
    \label{eq:transfer}
\end{equation}

$T(\q)$ determines the ultimate resolution and information limits of the microscope, and can be thought as describing the extent to which information from the sample scattering from different angles contributes to the information contained in the final measured image. $T(\q)$ decays at high frequencies and is oscillatory, which has traditionally complicated the interpretation of the contrast present in HRTEM micrographs. In spherical aberration-corrected microscopes, both ultimate resolution and information limits typically arise from high spatial frequency damping cause by chromatic aberration of the electron gun \citep{kirkland2020advanced, williamsTransmissionElectronMicroscopy2008}

\begin{equation}
    E(\q) = \exp( -(\pi \lambda \delta)^2 \norm{\q}^4 / 2)
    \label{eq:envelope}
\end{equation}

\noindent where $\delta$ is the defocus spread due to chromatic aberration. The general form of $E(\q)$ can be quite complex---e.g., by including effects of partial coherence \citep{spenceHighResolutionElectronMicroscopy2017}---but we will consider contributions to $E(\q)$ arising only from chromatic aberration.

While resolution limits arise mostly due to the damping envelope $E(\q)$, the qualitative form of $T(\q)$ and thus the behavior of information transfer from sample to image is instead dominated by the aberrations of the objective lens. The objective lens aberrations are captured in the aberration phase shift $\chi(\q)$, which, following \citep{rangeldacostaPrismatic20Simulation2021}, we define as 
    
\begin{multline}
    \chi(\q) = \sum_{m,n} (\lambda \norm{\q} )^m C^{\text{mag}}_{mn}  [ \cos{\left(n C_{\theta, mn}\right)}\cos{\left(n \q_\theta\right)} \\
    + \sin{\left(n C^{\theta}_{mn}\right)}\sin{\left(n \q_\theta\right)}].
\label{eq:chi_expansion}
\end{multline}

\noindent $\q_\theta$ is the polar angle of $\q$ as returned by the two-argument arctangent function, $\lambda$ is the electron wavelength, and $C^{\text{mag}}_{mn}$ and $C^{\theta}_{mn}$ are dimensionless coefficients corresponding to the aberration magnitude and azimuthal phase for an aberration of radial order $m$ and azimuthal order $n$. Eq. \ref{eq:chi_expansion} can be further reduced with the angle-sum trigonometric identity to

\begin{equation}
\label{eq:fast_ab_expansion}
    \chi(\q) = \sum_{m,n} \left[  (\lambda \norm{\q} )^m  C_{m,n}^{\text{mag} } \cos \left( n (C_{m,n}^{\text{ang}} - \q_\theta) \right) \right].
\end{equation}

We distinguish two sets of imaging conditions only by their lens aberrations and characterize their differences with respect to their corresponding information transfer properties via two auxiliary metrics, $\epsilon(\chi(\q))$ and $\sigma(\chi(\q), \chi'(\q))$, derived from $T(\q)$. Since it is usually unambiguous, for brevity we write $\epsilon(\chi) = \epsilon(\chi(\q))$ and $\sigma(\chi, \chi') = \sigma(\chi(\q)), \chi'(\q))$. We define the first, $\epsilon(\chi)$, as

\begin{equation}
    \epsilon\left(\chi\right) = \frac{\int \abs{T(\q)}^2 \mathrm{d}\q}{\int \abs{E(\q)}^2 \mathrm{d}\q}
    \label{eq:epsilon_T}
\end{equation}

\noindent which can be thought of as measuring the fraction of possible information that is transferred during imaging, as limited by high frequency damping. We note that we use $\abs{T(\q)}$ instead of $T(\q)$ to consider only the amount of information transferred at each frequency. The second metric, $\sigma(\chi, \chi')$, is defined as an overlap metric

\begin{equation}
    \sigma\left(\chi, \chi'\right) = \frac{\int \abs{T(\q)}\abs{T'(\q)}\mathrm{d}\q}{\int \abs{T(\q)}^2 \mathrm{d}\q}
    \label{eq:sigma_T}
\end{equation}

\noindent which can be understood as measuring the extent to which a new imaging condition, $\chi'$ (e.g., a test dataset), transfers information in similar amounts and at similar frequencies as the first, $\chi$ (e.g, a training dataset).  We normalize $\sigma(\chi, \chi')$ by the maximum possible overlap to the training dataset to be able to asymmetrically compare overlap conditions, which will be helpful in understanding the necessarily asymmetric relationships between training and test datasets in neural network training.

To demonstrate, in Fig. \ref{fig:emetric}a, we show how the total transferred information varies across imaging conditions for varying defocus and spherical aberration. When imaging with both defocus and spherical aberration close to zero, $\epsilon(\chi)$ is also close to zero---no information is transferred from the sample in an aberration-free lens---and when defocus and spherical aberration have the same sign, $\epsilon(\chi)$ has almost constant value. It becomes possible to form passbands in the contrast transfer function \citep{spenceHighResolutionElectronMicroscopy2017} when spherical aberration and defocus have opposite sign, and in turn, these passbands form local maxima of $\epsilon(\chi)$. We note here that these local maxima do not correspond to the widest passbands---those with the largest windows of interpretable image frequencies, such as Scherzer defocus---but are rather slightly periodically offset from such the sequence of passbands (Fig. \ref{fig:emetric}b).

\begin{figure}[t]
    \includegraphics[width=3.4in]{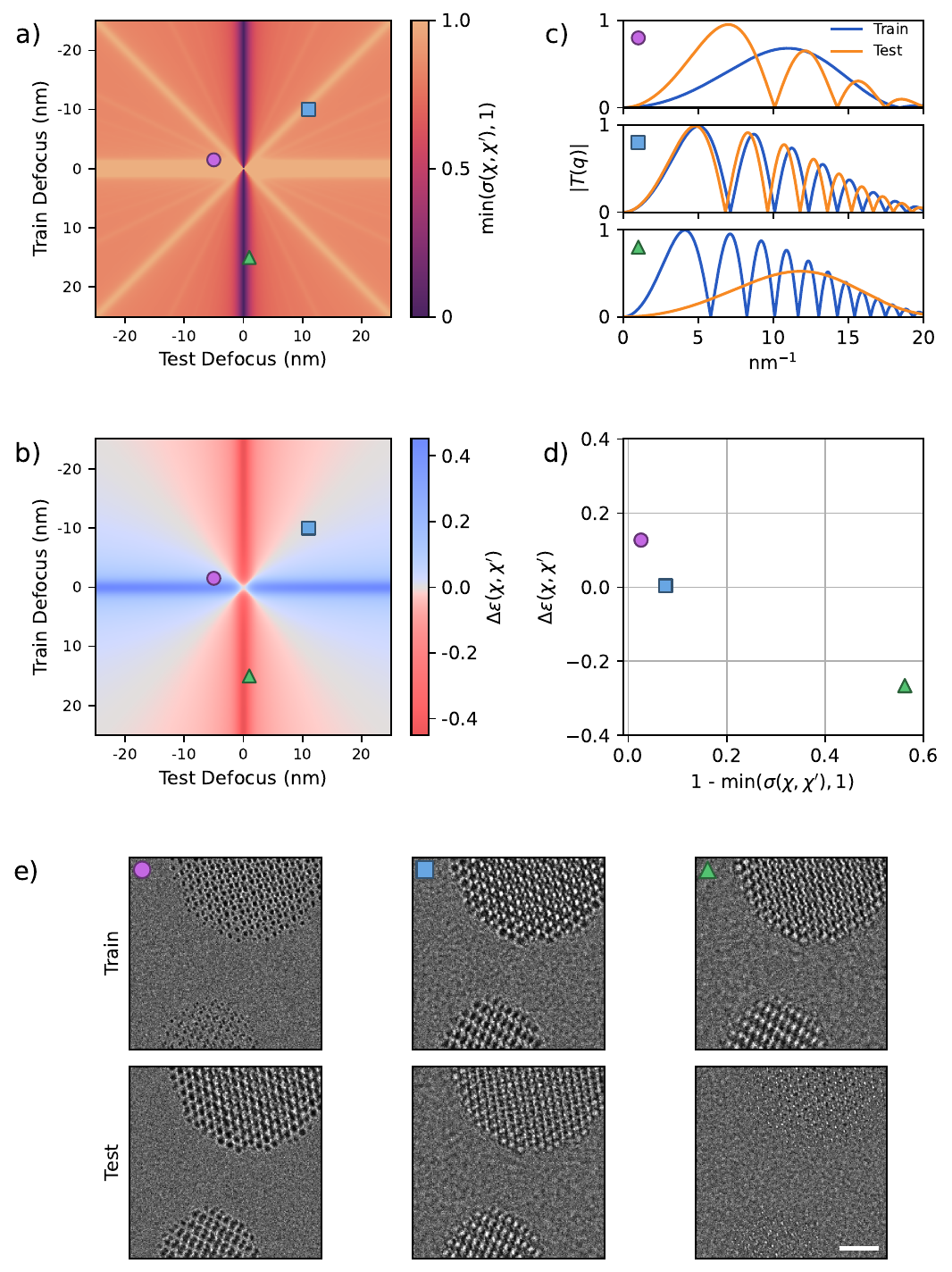}
    \caption{Comparative visualization of (a) $\sigma(\chi, \chi')$ and (b) $\Delta \epsilon(\chi, \chi')$ for pairs of imaging conditions differing only in defocus. (c) Selected pairs of absolute contrast transfer functions (CTFs) corresponding to defocus conditions indicated by circle, square, and triangle markers. (d) Scatter plot demonstration of how these transfer functions relate to each other via the metrics $\sigma(\chi, \chi')$ and $\Delta \epsilon(\chi, \chi')$. (e) Example image pairs of CdSe nanoparticles on amorphous carbon matching selected CTF pairs. Imaging conditions correspond to a TEM operated at 300kV with a focal spread of 10\r{A}. Scalebar is 1nm. }
    \label{fig:metric_explainer}
\end{figure}

In Fig. \ref{fig:metric_explainer}, we further show how $\epsilon(\chi)$ and $\sigma(\chi, \chi')$ together can be used to compare imaging conditions, considering a simple case in which two HRTEM datasets vary only in defocus. Generally, $\sigma(\chi, \chi')$ is maximized when information is transferred proportionally at similar frequencies, i.e., when the contrast transfer functions of two datasets closely align, which occurs here when the magnitudes of defocus are similar between train and test datasets (Fig. \ref{fig:metric_explainer}a,c). We use $\epsilon(\chi)$ to compare the total information content between HRTEM two datasets via their difference $\Delta\epsilon(\chi, \chi') = \epsilon(\chi') - \epsilon(\chi)$ (Fig. \ref{fig:metric_explainer}b), which we note is not sensitive to the spatial frequency dependence of the two transfer functions. A particularly extreme case can occur when either the training or testing dataset comes from near an aberration-free condition---in Fig. \ref{fig:metric_explainer}, these are conditions similar to that indicated by the triangle marker. When the training dataset nears zero defocus, the contrast transfer function decays in magnitude and loses its oscillations, causing $\sigma(\chi, \chi') \geq 1$ and further implying that the training dataset has strong overlap to nearly any other aberration condition. Conversely, $\sigma(\chi, \chi') \rightarrow 0$ when instead the test dataset nears zero defocus. Such conditions also cause $\Delta\epsilon(\chi, \chi')$ to be large in magnitude, indicating significant differences in the amount of total information available between two datasets. 

\subsection{Training protocol and hyperparameters}

All neural networks were optimized against a training dataset of 512 images before augmentation and a validation dataset of 128 images without augmentation.. Our neural network architecture follows previous work: we utilize a U-Net with a ResNet backbone \citep{rangeldacostaRobustSyntheticData2024, ronnebergerUNetConvolutionalNetworks2015},  containing approximately 14 million trainable parameters. A visualization of the model architecture is included in the supplementary material (Supplemental Fig. 1). We use a cross-entropy loss function for training and evaluation, and adjust loss values by the minimum attainable cross-entropy loss, corresponding to a perfect two-class prediction, so that all loss values fall between 0 and 1.

We trained two major sets of neural networks. Our first set of neural networks, comprising 3520 total models, focuses on understanding OOD generalization dynamics across different hyperparameter conditions. These models were optimized for 128 epochs with stochastic gradient descent using four different learning rates of 5e-4, 2.3e-3, 1.07e-2, and 5e-2, four different batch sizes of 8, 16, 24, and 32 images, and a constant learning rate decay factor of 0.98. For each hyperparameter condition, we train 20 models. Throughout training, neural network performance was measured on a series of CdSe defocus datasets, described in the following section. 

To further probe OOD behavior with respect to perturbations in imaging conditions, we train and evaluate a second series of models against training datasets with randomly sampled imaging conditions, this time with fixed hyperparameter conditions. The models were again trained with stochastic gradient descent with a learning rate of 1e-2, a batch size of 16, and a constant learning rate decay factor of 0.99. Models were optimized for up to a maximum of 32 epochs; we saved  model weights at their best validation loss and after passing validation losses of 0.1 and 0.05, after which we conclude training. An additional 4096 models were trained for these series; 3745 models passed the 0.1 loss threshold and 3089 models passed the 0.05 loss threshold. Only a small number of models surpassed the 0.025 loss threshold. After training, models were evaluated on their OOD performance against up to 32 randomly sampled evaluation datasets, which, after accounting for training and evaluation failures (primarily due to computation time outs), resulted in 80896 and 72864 total evaluations for the 0.1 loss 0.05 loss models, respectively.

\subsection{Dataset Generation and Descriptions}

Our segmentation dataset generation pipeline follows previous work \citep{rangeldacostaRobustSyntheticData2024}. Throughout the structure and image generation process, we randomly sample structural configurations and imaging conditions. Unless specified otherwise, the sampling of parameters should be understood to be performed with uniform sampling throughout the range of a parameter. We partition training and validation dataset splits by structure, i.e., all models, per series, were trained on a single set of common structures, and validation performance (used to quantify both ID and OOD performance) was measured with a unique,  disjoint set of structures. 

We generate structures of nanoparticles on substrates using the Construction Zone software package and perform two simulations for each: one of the full structure, and one of just the nanoparticles in vacuum. Each simulation structure contains between one and five spherical nanoparticles with radii between 1 and 2.5nm, separated by a distance of at least 1nm. We embed each particle atop an amorphous carbon substrate \citep{rangeldacostaRobustSyntheticData2024, ricolleauRandomVsRealistic2013} and at uniformly random orientations; the substrates have 12.8nm square areas and thicknesses between 2 and 8nm. The carbons substrate generation is further constrained to also satisfy periodic boundary conditions in the planar directions to prevent boundary artifacts in the TEM simulation. For our datasets containing only wurtzite CdSe nanoparticles, we add between 0 and 1 stacking faults to particles larger than 3nm in diameter and between 0 and 2 stacking faults to particles larger than 4.4nm; we describe the generation of non-CdSe structures in Supplementary Section 2.

HRTEM simulations were performed with the multislice algorithm \citep{rangeldacostaPrismatic20Simulation2021, abtem}, using a potential sampling of 0.1\AA, a slice thickness of 2\AA, eight frozen phonons, and infinitely projected potentials with the Kirkland potential parameterization \citep{kirkland2020advanced}. The final images are generated from the full structure simulations, while the segmentation labels are calculated from the electron wavefunction phase of the nanoparticles-in-vacuum simulation. We calibrate zero defocus to the minimum contrast condition of the nanoparticles in vacuum, which we heuristically determine via a two stage grid search to within 0.1\AA. After application of imaging conditions and noise effects, we randomly crop images into patches of 512 by 512 pixels and then generate image augmentations using all eight possible dihedral augmentations arising from 90 degree rotations and flips along horizontal and vertical axes.

\begin{table}[]
    \centering
    \begin{tabular}{lrr}
    \hline
     Aberration & Coefficient & Sampling Magnitude \\
     \hline
     Defocus & $C_{20}$ &  15 nm \\
     Two-fold astigmatism & $C_{22}$ & 3 nm \\
     Axial coma & $C_{31}$  & 0.2 $\mu$m \\
     Three-fold astigmatism & $C_{33}$ & 0.2 $\mu$m \\
     Spherical & $C_{40}$ & 0.1 mm \\
     \hline
    \end{tabular}
     \caption{Maximum magnitudes of sampled aberration coefficients for synthetically generated HRTEM images.}
     \label{tab:aberrations}
\end{table}

We make two main sets of training datasets, one varying only in the image defocus, and the other including aberrations up to spherical aberration. For all images, we apply an electron dose and focal spread of 300e-/\AA$^2$ and 10\AA, respectively. The defocus datasets consisted of imaging conditions sampled around focal points from -25nm to +25nm in steps of 5nm and with a standard deviation of $\pm$1nm.

For our second series of datasets, we sample imaging conditions and generate image batches on-the-fly during model training and evaluation. For each individual image batch, we sample a target imaging condition by uniformly sampling aberration magnitudes up to a set of maximum values, given in Table \ref{tab:aberrations}. Within each image batch, different images will have slightly different imaging aberrations, too. We sample aberrations for each image from Gaussian distributions with means given by the previously sampled target imaging condition and the variances given as 1/8th of the maximum magnitudes. Aberration rotations were sampled with magnitudes up to 0.125 radians and variances of $\pi/16$ radians. 

Within the second series of datasets, we additionally sample training dataset aberrations along a fixed combinations of defocus and spherical aberration to improve the dispersion of sampled aberration pairs with respect to $\epsilon$ and $\sigma$. For these aberrations, we fix defocus and spherical aberration via the following equation

\begin{equation}
    C_{20} = -\sqrt{n \lambda C_{40} }
    \label{eq:passband}
\end{equation}

\noindent where $\lambda$ is the electron wavelength. We define $n$ as the passband order (differing in notation from the standard formulation used to analytically determine the Scherzer focus condition \citep{spenceHighResolutionElectronMicroscopy2017}). We sample 64 target aberrations along eight different passbands with $n= k/4$ for $k$ from 1 through 6, 9, and 13. The 64 spherical aberration coefficients are linearly spaced between $0.1\, \mu\text{m} \leq C_{40} \leq C_{40, \text{max}}$. We adjust the maximum magnitude of $C_{40}$ so that the magnitudes of both defocus and spherical aberration remain smaller than the magnitudes indicated in Table \ref{tab:aberrations}, setting $C_{40,\text{max}} =\min \left( 0.1\text{mm}, (C_{20}^2 / (\lambda n) \right)$. As before, we sample the remaining aberrations uniformly at random but with reduced maximum magnitudes, scaled by a multiplicative factor of 0.2.

\section{Results}

\subsection{Segmentation network training and generalization dynamics across imaging defoci}

We begin by investigating the out-of-distribution (OOD) generalization behavior solely with respect to image defocus in HRTEM, a relatively constrained setting which is nonetheless challenging to assess experimentally due to difficulties in precisely measuring and calibrating defocus. Moderate shifts in defocus can induce strong, nonlinear contrast effects which significantly alter image textures. At especially large defocus, image delocalization effects \citep{williamsTransmissionElectronMicroscopy2008} and rapid oscillations of the contrast function make HRTEM micrographs especially hard to interpret and precisely measuring nanoparticle boundaries difficult. Altogether, these effects can impact the performance of neural networks used to segment atomic resolution HRTEM images. To help visually demonstrate how model performance degrades, and to provide some quantitative intuition for how different magnitudes of cross-entropy loss relate to final segmentation quality, we include example images of model segmentation throughout various stages of training in Supplemental Fig. 2.

\begin{figure}[!t]
    \centering
    \includegraphics[width=3.4in]{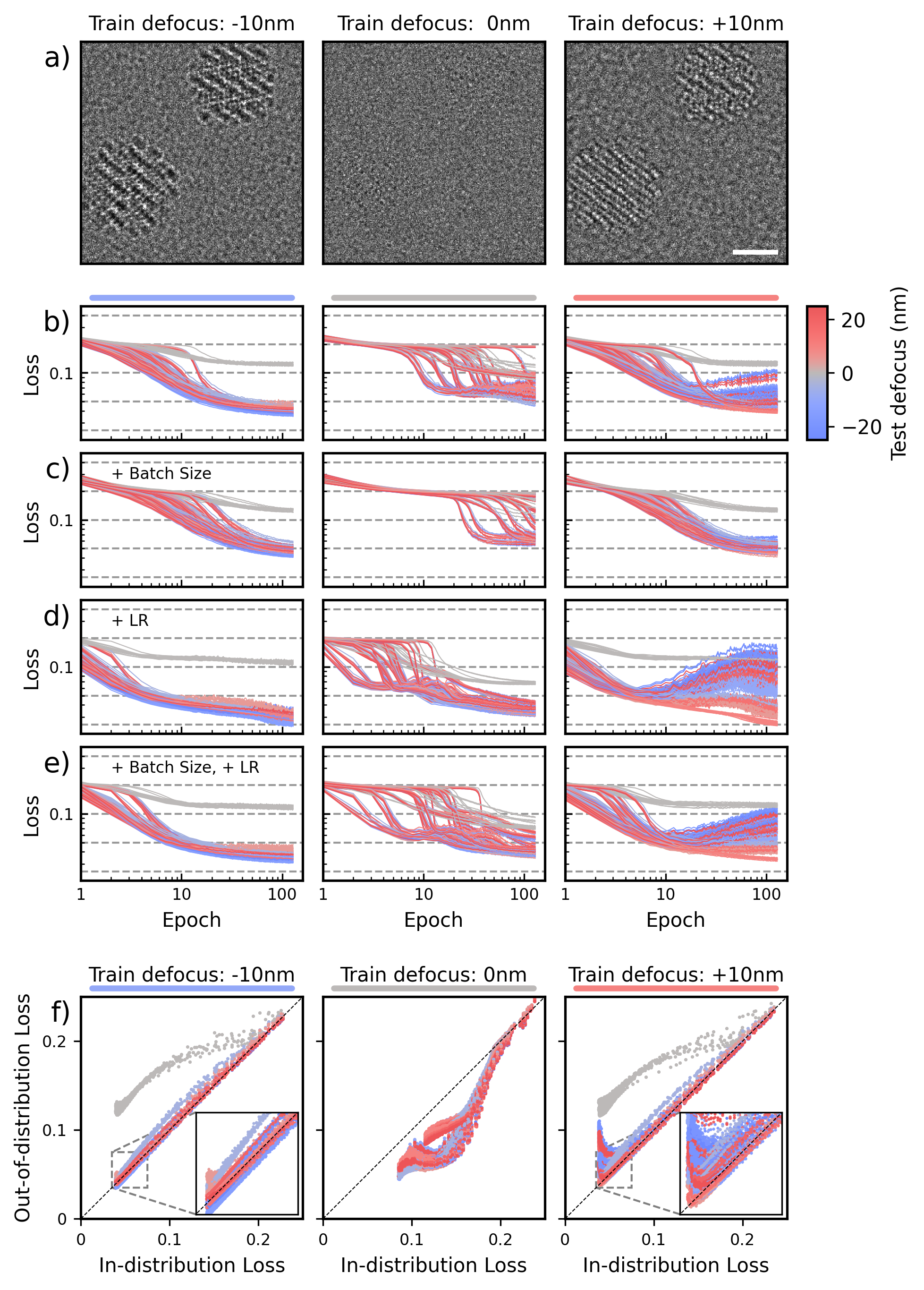}
    \caption{Out-of-distribution generalization behavior throughout training of U-Net image segmentation models trained on -10nm (left column), 0nm (middle column) and +10nm (right column) defocus images of CdSe nanoparticles. Defocus is measured from the minimum contrast condition of the CdSe nanoparticles. (a) Example regions from a simulated micrograph at each corresponding defocus for visual reference; scale bar is 1nm. (b--e) Learning curves for series of models trained with learning rates of 2.3e-3, 2.3e-3, 1e-2, 1e-2 and alternating batch sizes of 16, 32, 16, 32; line color corresponds to nominal defocus of validation dataset. (f) Correlation between in- and out-of-distribution losses for models trained with learning rates of 1e-2 and batch size of 16 (d), aggregated across all training times.}
    \label{fig:dynamics}
\end{figure}

\begin{figure}[t!]
    \centering
    \includegraphics[width=3.4in]{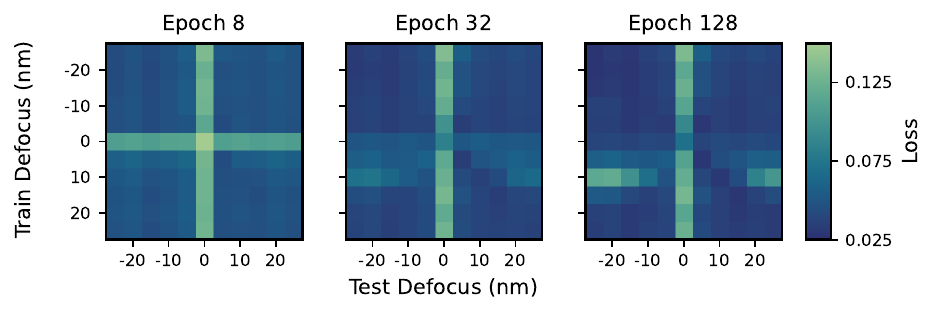}
    \caption{Confusion matrix visualization of neural network loss as trained and evaluated on CdSe image datasets simulated with mean defoci between -25 and +25nm at 300kV and with a defocus spread of 10 \r{A}. From left to right, we visualize model performance at epochs 8, 32, and 128. Visualized loss is taken as mean of twenty models trained with a learning rate of 1e-2 and batch size of 24. Smaller loss is better.}
    \label{fig:defocus_gen}
\end{figure}

Utilizing a synthetic data-generation framework, we generate large simulated datasets of HRTEM images of CdSe nanoparticles on amorphous carbon substrates and train an ensemble of neural network models on datasets with defocus ranging from -25nm to +25nm in steps of 5nm. We calibrate defocus to the minimum contrast condition of the nanoparticles. In total, across the eleven defocus conditions, we train 3520 different models using 16 different hyperparameter settings, training 20 different randomly initialized models for each configuration to fully capture any distributional performance behavior, and at each epoch in training, we evaluate neural network performance on all defocus datasets. In Fig. \ref{fig:dynamics}, we visualize ID (validation) and OOD training curves for models trained on -10nm, 0nm, and +10nm defocus datasets across a selection of hyperparameter configurations and compare the trends and dynamics of neural network generalization. OOD performance tends to correlate with in-distribution performance (Fig. \ref{fig:dynamics}f), even though OOD performance is often slightly worse throughout training. The extent of performance degradation seems to depend slightly on the relative shift in defocus magnitude. In general, the 0nm defocus dataset appears to be significantly more difficult for the neural networks to predict and learn from, as indicated by overall slow convergence rates of the 0nm training curves and the high loss at the end of training. However, almost universally, the 0nm-trained models are more successful out-of-distribution than they are in-distribution. For the +10nm-trained models, there appears to be a significant onset of overfitting as indicated by an increase in loss and variance in performance at later stages of training (c.f. Fig. \ref{fig:dynamics}b,d,e, right). 
    
The choice of training hyperparameters does not seem to significantly affect absolute performance or generalization behavior, but do seem to impact the training dynamics within a fixed training budget. Increasing the learning rate causes all models to reach their asymptotic optima earlier (c.f. Fig. \ref{fig:dynamics}b,d,e), but also leads to the models training on the +10nm defocus dataset to enter deeper into an overfitting regime. Alternatively, increasing the batch size slightly slows down model convergence (c.f. Fig. \ref{fig:dynamics}b,c,e) but, for the non-zero defocus datasets, seems to reduce the variance in model performance and lessen the degree of overfitting. For the 0nm-trained models, there is significantly higher variance on the dynamics of training, which may be an effect of decreased training stability due to the difficulty of the training dataset. Generalization dynamics trends are qualitatively similar across the remaining training defoci, for which we include results in Supplemental Figs. 3 and 4.

To further disentangle the relationship between training dataset defocus and neural network OOD performance across all defoci, we take a more detailed view of a single hyperparameter configuration of our models and explore performance across a series of training snapshots. We focus on three training stages---early in training, before models reach asymptotic convergence; near the optimum, as models start converging; and late in training, when models may begin overfitting. Using confusion matrices, in Fig. \ref{fig:defocus_gen} we visualize mean neural network OOD performance across all networks trained with a batch size of 24 images and a learning rate of 1e-2 (corresponding to Fig. \ref{fig:dynamics}e). We highlight performance after the 8th, 32nd, and 128th (final) training epochs, corresponding to the three training stage described prior. Model performance during early phases of training seems relatively homogeneous across defocus for all models and performance out-of-distribution does not notably differ from in-distribution, except, again, for models trained with the 'in-focus' 0nm defocus dataset. After model performance begins to converge, OOD performance qualitatively settles, and may get slightly worse as training continues to progress, which can be seen when comparing performance of models trained on +5nm and +10nm defocus at epochs 32 and 128. 

Model performance is best OOD when models evaluate test datasets containing images captured with similar defoci. Surprisingly, given the drastic image changes which can occur from contrast reversal effects, our models still perform about as well (but slightly worse, c.f. Supplemental Figure 5) on defoci similar in magnitude but with opposite sign. Importantly, there is a significant asymmetry in the OOD performance trends between the training and test dataset pairs. When not overfitting, model performance on the 0nm defocus dataset is usually the worst out of all ID and OOD datasets, across both different training defoci and training stages, improving only slightly during later stages in training. Yet, when training on the 0nm defocus data, the models have the best OOD generalization---managing to perform better OOD than ID---while failing fail to fully converge during training. Models trained on slightly positive defocus had the worst generalization and most susceptibility to overfitting, which might be due to the intrinsic asymmetry of the atomic structures with respect to the electron beam---as simulated, the nanoparticles are always the first structure with which the electron beam interacts. 

\subsection{Use of contrast transfer functions to characterize out-of-distribution generalization}

The performance across defocus, and especially the performance on the 0nm defocus datasets, can be partially understood by considering the relationship between contrast transfer functions (CTFs) of the two datasets. We introduce two auxiliary metrics, detailed in the Methods section, for comparing different imaging conditions by their CTFs: $\epsilon(\chi)$, a measure which could be interpreted as the amount of information transferred under a certain imaging condition, and $\sigma(\chi, \chi')$, which is an asymmetric measure of the relative information transfer overlap between two imaging conditions. The auxiliary metrics $\sigma(\chi, \chi')$ and $\epsilon(\chi)$ let us readily expand our analysis beyond the simple case of varying just defocus to any arbitrary sets of imaging conditions. 

\begin{figure}[t]
    \centering
    \includegraphics[width=3.2in]{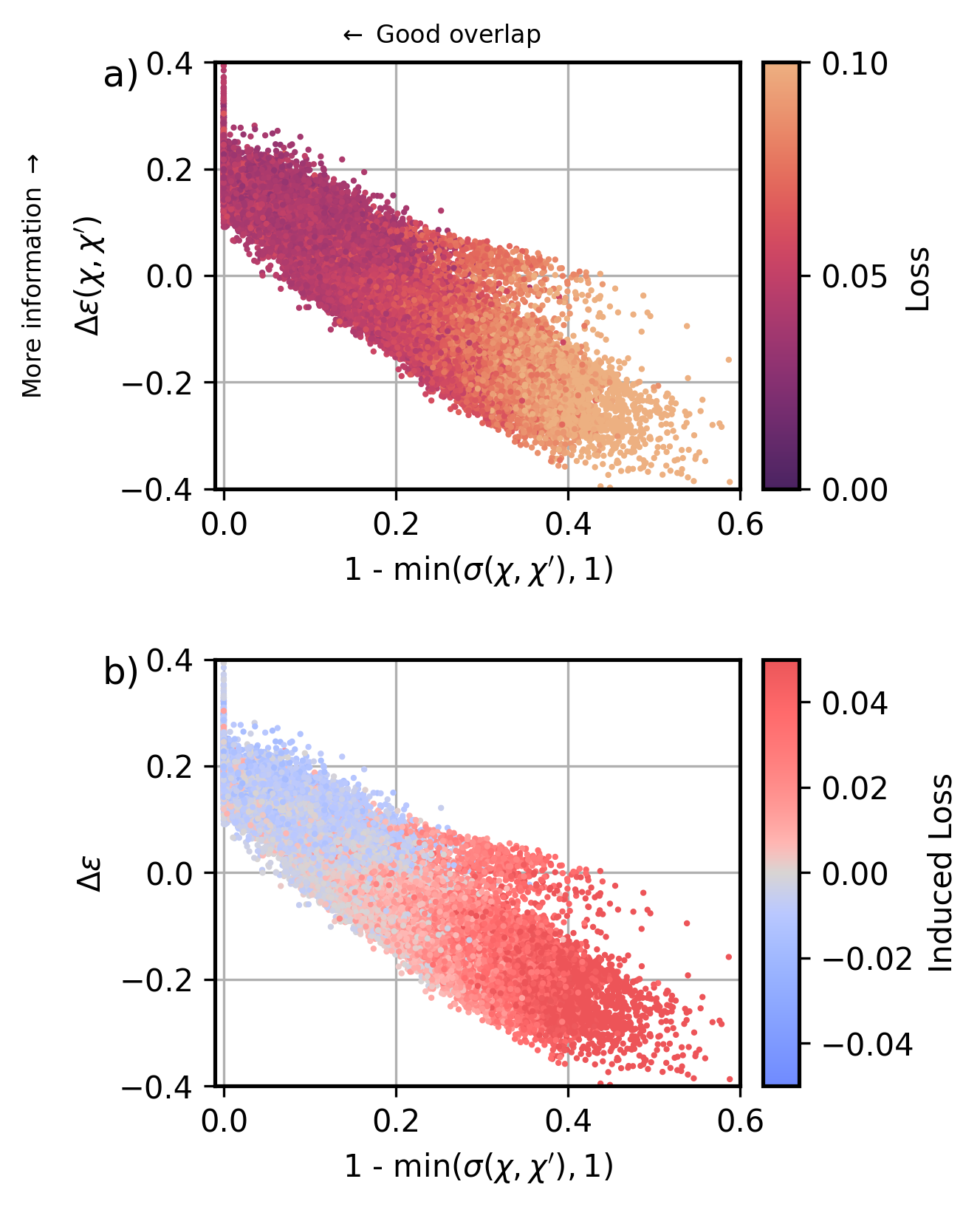}
    \caption{Neural network loss (a) and induced (excess) loss over the training dataset (b) out of distribution, relative to $\sigma(\chi, \chi')$ and $\Delta \epsilon(\chi, \chi')$. Relative information content increases vertically, and information overlap decreases to the right. More negative induced loss (blue) indicates a model improving OOD, whereas a more positive induced loss (red) indicates the model worsening. OOD evaluations were performed using randomly generated model training datasets and evaluation datasets. Aberrations of training and evaluation datasets were sampled up to 5th order (defocus, axial coma, two- and three-fold astigmatism, and spherical), and models were trained until reaching a loss threshold of 0.05. Each marker corresponds to a unique training dataset--OOD dataset aberration pair, in total encompassing 3089 unique models and 72864 unique dataset pairs.}
    \label{fig:es}
\end{figure}

Under this framework, we have performed another large ensemble neural network training series, training over 3000 new models on randomly sampled imaging conditions with up to fifth-order aberrations. The models are trained for a fixed training period, and we save model weights at the optimal training epoch, as well as after models have passed loss thresholds of 0.1 and 0.05. After training, for each model and weight configuration, we sample 32 new imaging conditions and evaluate the model performance out-of-distribution, including over 72,000 unique model OOD evaluations.  In Fig. \ref{fig:es} we organize the OOD evaluations by their relative change in information quantity ($\Delta \epsilon(\chi, \chi')$) and overlap ($\sigma(\chi, \chi')$) between the training dataset and test datasets. Model loss OOD smoothly varies across differences in information content and overlap (Fig. \ref{fig:es}a). When $\sigma(\chi, \chi')$ is close to 1---indicating strong overlap to the OOD imaging conditions---and $\Delta \epsilon(\chi, \chi')$ is small, neural network OOD performance does not differ strongly from ID performance. For large, positive $\Delta \epsilon(\chi, \chi')$, i.e., when the OOD dataset has more information transferred from the nanoparticles to the final image, the models can even perform better OOD than on their own training data (Fig. \ref{fig:es}b). On the other hand, when $\Delta \epsilon(\chi, \chi')$ is large and negative, or when $\sigma(\chi, \chi')$ is small, OOD performance suffers, worsening as $\sigma(\chi, \chi')$ continues to decrease. Naturally, large negative values of $\Delta\epsilon$ can only occur when $\epsilon(\chi)$ itself is large, i.e. when the training dataset imaging conditions impart a significant amount of information transfer. Such datasets are in fact demonstrably easier to learn from: absolute model convergence and training convergence rates on in-distribution data both reliably improve with increasing $\epsilon(\chi)$ (c.f. Supplemental Fig. 6) indicating that---perhaps intuitively---neural network training dynamics strongly depend on the amount of information available in training datasets. Overall, there appears to be a zone of stability with large overlap and positive $\Delta \epsilon(\chi, \chi')$ where the neural network models perform comparatively well out-of-distribution, relative to their performance in-distribution.

\section{Discussion}

Our main result presented in Fig. \ref{fig:es} indicates that, for HRTEM image segmentation tasks, neural networks robustly handle small shifts in imaging conditions. Under larger perturbations to experimental imaging conditions, neural networks seem to gracefully and smoothly fail, and the extent of their failure can be directly understood by considering changes in the contrast transfer function of the microscope. Generally, this gives us confidence that neural network models for TEM analysis can be safely and meaningfully deployed in environments with (relatively controlled) shifting imaging conditions, and that we can rely on well-developed theories of imaging in electron microscopy to guide future development of machine learning approaches. Moreover, our results also imply that, with respect to imaging conditions, training datasets can be designed to circumvent failures in OOD generalization. In practice, one can avoid degradation of performance by constructing training datasets which span diverse information transfer conditions and thus ensuring that any arbitrary OOD imaging condition has strong information overlap to some subset of training data.

For practical deployment of neural networks in electron microscopy, it is prudent to consider limitations and possible extensions of our information transfer framework. We believe our approach can be readily adapted to other TEM imaging modalities, but it is not immediately clear that it would be appropriate across all types of machine learning models and tasks, e.g., regression with known differences in known OOD behavior \citep{ahujaInvariancePrincipleMeets2022}. OOD generalization behavior for supervised learning for classification might behave similarly given the similarities in construction between classification and segmentation tasks\citep{hooper2023a}.

When training neural network models for TEM analysis with simulated data, it is particularly important to understand any remaining gaps between simulated training data and experimental test data. For TEM applications, these emulation gaps largely involve describing and capturing representative imaging conditions and accurately accounting for the distribution of possible atomic structures. In this work, we only consider shifts in the imaging conditions due to shifts in the aberration conditions and high-frequency damping from chromatic aberration. We expect that detector effects and their corresponding modulation transfer functions, which are simple to model and incorporate into HRTEM simulation \citep{leeElectronDoseDependence2014a, madsenDeepLearningApproach2018a} but more difficult to measure, could directly be incorporated into Eq. $\ref{eq:transfer}$ as another multiplicative term. A fuller treatment of the envelope function (c.f. Eq. \ref{eq:envelope}) can be performed by incorporating the contributions from the partial coherence of the source, sample drift and vibration, and the detector \citep{spenceHighResolutionElectronMicroscopy2017, williamsTransmissionElectronMicroscopy2008}. No further alterations to the metrics $\epsilon(\chi)$ and $\sigma(\chi, \chi')$ need to be made when experiments differ in their envelope functions. However, while differences in $\epsilon(\chi)$ remain bounded between -1 and 1 when multiple envelope functions are considered, the relative magnitudes of $\Delta \epsilon(\chi, \chi')$ and $\sigma(\chi,\chi')$ would shift, e.g., when the envelopes differ significantly in their high frequency cutoffs, which might make consistent interpretation of the two metrics across all imaging conditions more difficult. 

Shifts in the distribution of the atomic structures are significantly more difficult to describe and anticipate in experiments---and, perhaps, more pressing to consider. To possibly incorporate structural information into our framework for predicting OOD performance under shifts of atomic structure distributions, we can introduce weighted versions of the transfer function in Eqs. \ref{eq:epsilon_T} and \ref{eq:sigma_T} by substituting $T(\q)$ with $\Tilde{T}(\q) = \omega(\q) T(\q)$, where $\omega(\q)$ is some spectral weighting function calculated from the atomic structures, e.g., arising from the atomic structure factors. We ran additional training experiments in which we trained segmentation networks on nanoparticle lattice systems varying in chemistry and symmetry (c.f. Supplemental Section 2), and show that OOD performance does not smoothly nor consistently vary across different regions of information overlap and information differences when using spectral weights determined derived from lattice structure factors (Supplemental Fig. 7), even when controlling for differences in chemistry. Thus, shifts in atomic structure do not seem to be simply explained within our information transfer model and more work is necessary to understand how shifts in atomic structures impact OOD model performance in TEM analysis tasks, and more broadly, to understand more fully how to describe shifts in atomic structure distributions. Recent work on describing the relative information content of atomistic training datasets for machine learning of interatomic potentials \citep{schwalbe-kodaModelfreeEstimationCompleteness2025} could be particularly pertinent, but did not yield meaningful results in our testing.

We strongly emphasize that our study relies on a fairly precise description of our data domain and of the OOD shifts we consider, using information theoretic approaches specific to electron microscopy. As we tackle more complex tasks on more challenging data with machine learning, producing such precise descriptions may become increasingly difficult. Domain agnostic methods for understanding OOD generalization behavior could be especially useful in circumventing challenges in describing differences in the relative information content between datasets for more complex tasks and datasets. Such methods exist, e.g., for detecting OOD training during model deployment \citep{madrasDETECTINGEXTRAPOLATIONLOCAL2020, sharmaSketchingCurvatureEfficient2021, liGeneralizedFewShotOutofDistribution2025, leeSimpleUnifiedFramework, haroushStatisticalFrameworkEfficient2022} or for conditioning models to respond more stably to OOD shifts from their training domains  \citep{rameDiverseWeightAveraging2023, rameFishrInvariantGradient2022, mahajanDomainGeneralizationUsing2021, arjovskyInvariantRiskMinimization2020}. However, few, if any, domain-agnostic approaches exist which can successfully calibrates or predict OOD generalization behavior; the authors are not aware of such an approach. Here, we explore the utility of second-order (Hessian) information of the model, known to strongly relate to the Fisher information \citep{bottcherVisualizingHighdimensionalLoss2024, rameFishrInvariantGradient2022}, and empirical measures of the Kullback-Leiber (KL) divergence between imaging datasets as two potential complementary and domain-agnostic approaches for quantitatively understanding the OOD behavior of neural networks. For datasets for which neural networks perform well OOD, the OOD eigenspectrum distributions of neural network loss Hessians evolve similarly to those of the training dataset (Supplemental Fig. 9). Similarly, the dominant subspaces of the loss Hessians align over the course of model optimization (Supplemental Fig. 10). Together, these indicate that loss landscape curvature could be a good predictive measure of OOD performance, and more fundamentally, that understanding the dependence of the loss landscape across different data distributions could provide deeper insight to OOD generalization behavior. From the data perspective, large KL divergences between datasets, which signify large differences in information content, moderately correlates to larger shifts in performance OOD, but it seems difficult to predict whether a model will degrade or improve (Supplemental Fig. 8). For more details on these experiments, we refer the reader to Supplemental Sections 3 and 4. Further investigation of accurate, inexpensive approaches for measuring domain shift and predicting OOD generalization can help bridge critical gaps in our understanding of generalizable machine learning approaches.

\section{Conclusion}

To summarize, our results indicate that neural networks for HRTEM analysis enjoy meaningful and stable OOD generalization behavior with respect to shifts in imaging conditions. Using a data-centric perspective and domain-specific information theoretic models, we can predict in which scenarios our models might begin to fail in experimental environments and can understand how to effectively design our training workflows to avoid such failures. Significant challenges remain, however, in our ability to fully describe our data at a distributional level, especially regarding atomic structures, which could remain a major barrier for understanding OOD generalization phenomena in materials characterization tasks. Still, we believe that maintaining a perspective focused on understanding how information arises in our experimental data and how machine learning models make use of such information will be crucial for refining our understanding and use of machine learning tools in applied scientific settings.

\section{Competing interests}
No competing interest is declared.

\section{Author contributions statement}

L.R.D. conducted the experiments and analyzed the results; L.R.D. and M.C.S. wrote and reviewed the manuscript; and M.C.S. supervised the work and acquired the funding.

\section{Acknowledgments}
Work at the Molecular Foundry was supported by the Office of Science, Office of Basic Energy Sciences, of the U.S. Department of Energy under Contract No. DE-AC02-05CH11231. This research used resources of the National Energy Research Scientific Computing Center (NERSC), a Department of Energy User Facility using NERSC award BES-ERCAP 32753. This material is based upon work supported by the U.S. Department of Energy, Office of Science, Office of Advanced Scientific Computing Research, Department of Energy Computational Science Graduate Fellowship under Award Number DE-SC0021110.

\bibliographystyle{unsrtnat}
\bibliography{refs}
\end{document}


\title{Supplementary Material: Contrast transfer functions help quantify neural network out-of-distribution generalization in HRTEM}

\author[1,2,*]{Luis Rangel DaCosta}
\author[1,2,*]{M.C. Scott}
\affil[1]{Department of Materials Science and Engineering, University of California Berkeley, Berkeley, CA 94720}
\affil[2]{National Center for Electron Microscopy, Molecular Foundry, Lawrence Berkeley National Laboratory, 1 Cyclotron Road, Berkeley, CA, USA}
\affil[*]{lrangeldacosta@lbl.gov, mary.scott@berkeley.edu}

\date{}
\maketitle

\maketitle

\section{Neural network architecture and additional performance measurements}

\begin{figure}[phtb]
\includegraphics[width=0.98\linewidth]{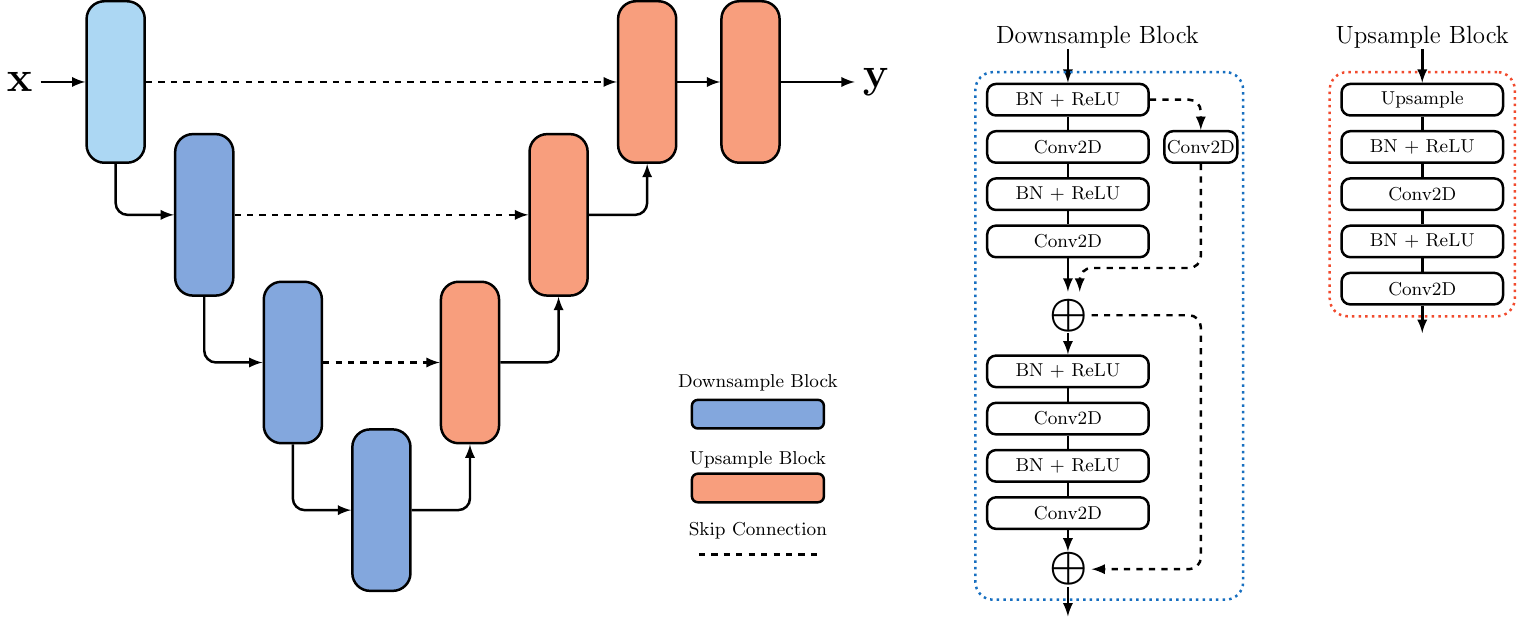}
\caption{U-Net model architecture employed in this study. In total, the model contains ~14 million trainable parameters. The initial downsampling block (light blue) consists only of a single batch normalization and a max pooling layer. }
\label{sfig:nn_arch}
\end{figure}

\begin{figure}
    \centering
    \includegraphics[width=\linewidth]{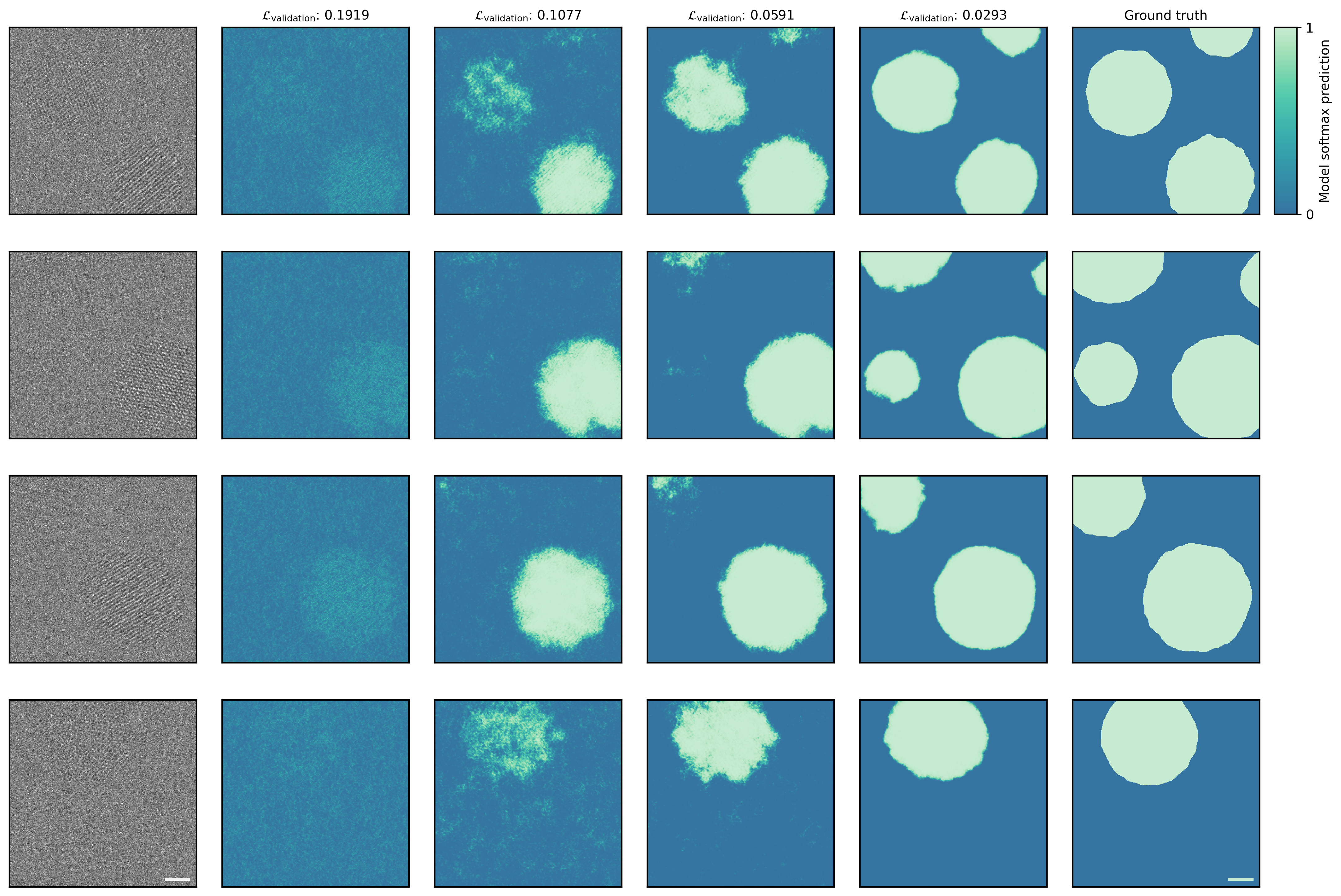}
    \caption{Example segmentation performance of neural network throughout different stages in optimization on CdSe nanoparticles with randomly sampled imaging conditions. Model improves in performance from left to right.}
    \label{sfig:examples}
\end{figure}

\begin{figure}
    \centering
    \includegraphics[width=\linewidth]{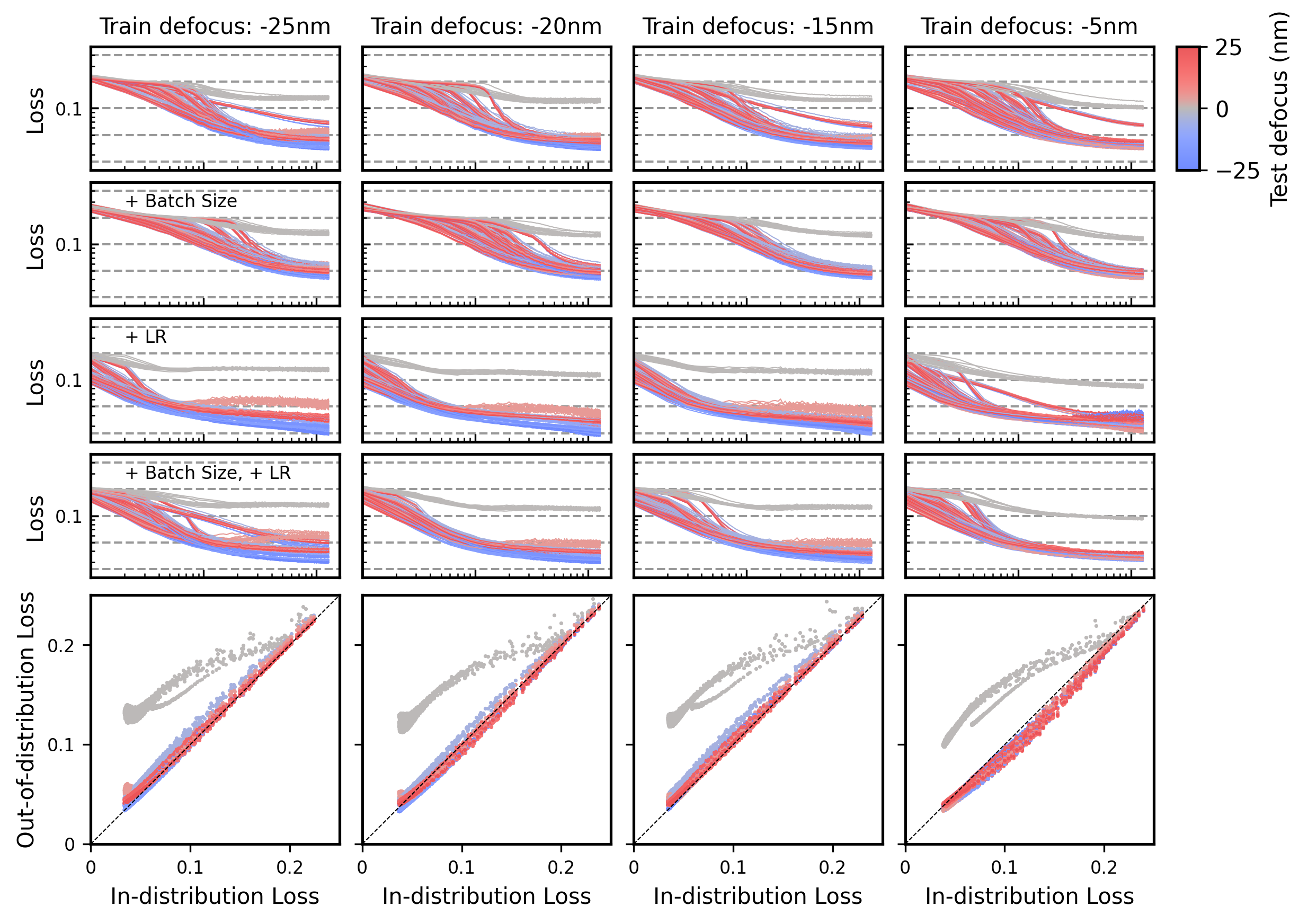}
    \caption{Out-of-distribution generalization behavior throughout training of U-Net image segmentation models trained on -25nm, -20nm, -15nm, and -5nm defocus images of CdSe nanoparticles, from left to right. Defocus is measured from the minimum contrast condition of the CdSe nanoparticles. From top to bottom, learning curves for series of models trained with learning rates of 2.3e-3, 2.3e-3, 1e-2, 1e-2 and alternating batch sizes of 16, 32, 16, 32; line color corresponds to nominal defocus of validation dataset. (bottom row) Correlation between in- and out-of-distribution losses for models trained with learning rates of 1e-2 and batch size of 16, aggregated across all training times.}
    \label{sfig:neg_dynamics}
\end{figure}

\begin{figure}
    \centering
    \includegraphics[width=\linewidth]{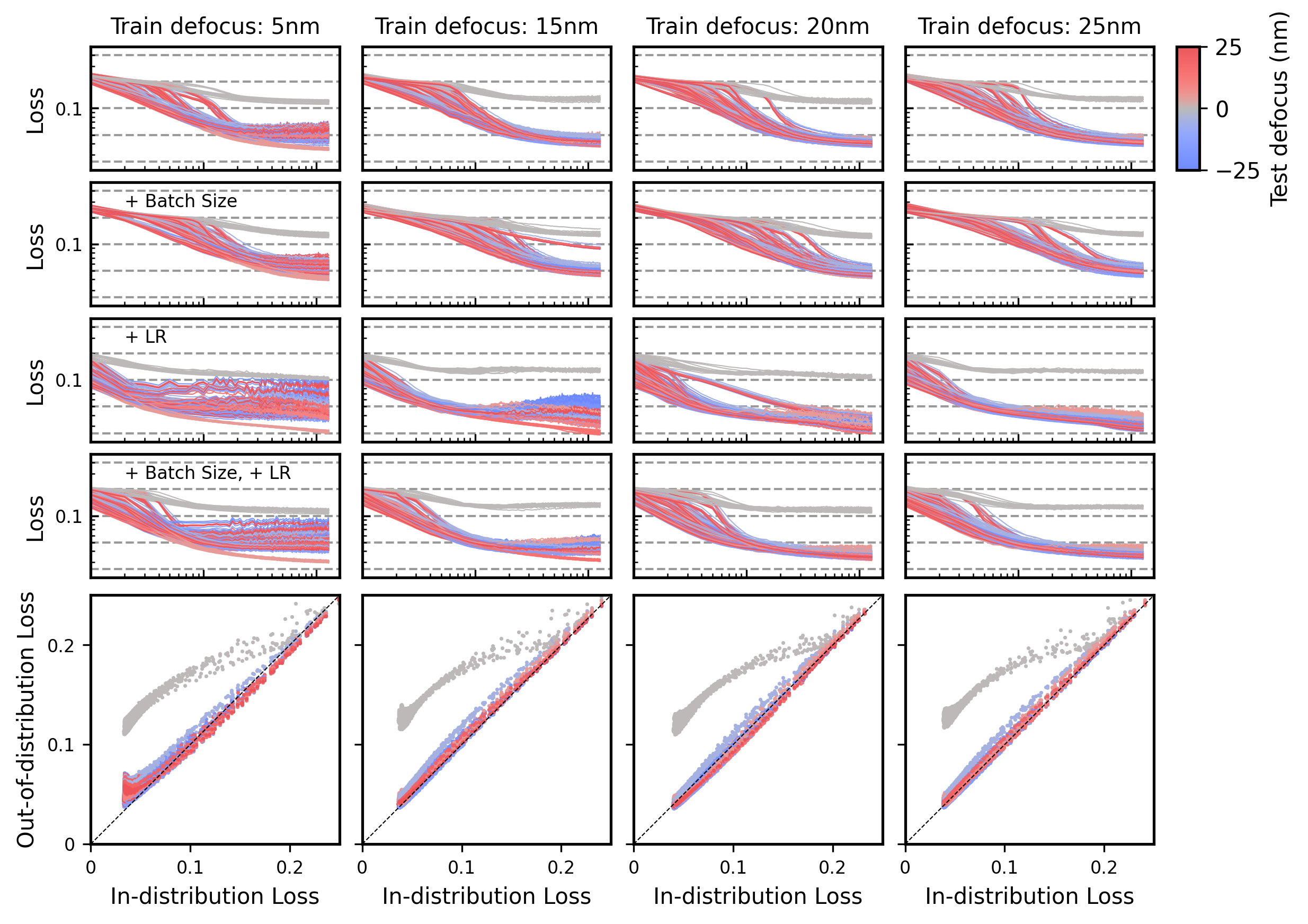}
    \caption{Out-of-distribution generalization behavior throughout training of U-Net image segmentation models trained on 5nm, 15nm, 20nm, and 25nm defocus images of CdSe nanoparticles, from left to right. Defocus is measured from the minimum contrast condition of the CdSe nanoparticles. From top to bottom, learning curves for series of models trained with learning rates of 2.3e-3, 2.3e-3, 1e-2, 1e-2 and alternating batch sizes of 16, 32, 16, 32; line color corresponds to nominal defocus of validation dataset. (bottom row) Correlation between in- and out-of-distribution losses for models trained with learning rates of 1e-2 and batch size of 16, aggregated across all training times.}
    \label{sfig:pos_dynamics}
\end{figure}

\begin{figure}[hbtp]
    \centering
    \makebox[\textwidth][c]{\includegraphics[width=4.8in]{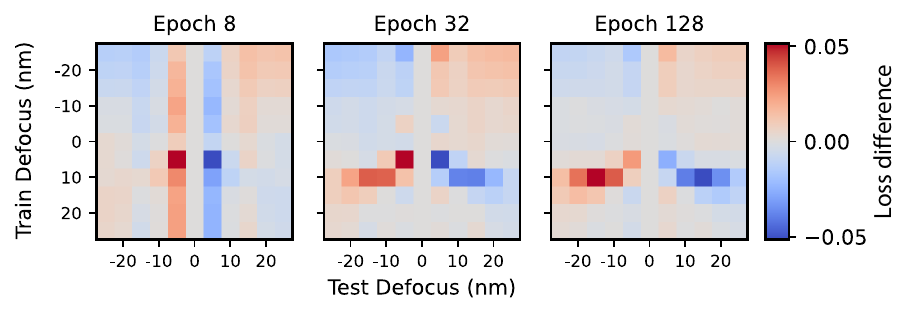}}
    \caption{Difference in neural network performance on defocus of opposite sign for neural networks trained and evaluated on CdSe image datasets simulated with mean defoci between -25 and +25nm at 300kV and with a defocus spread of 10 \r{A}.}
    \label{sfig:defocus_diff_map}
\end{figure}

\begin{figure}[h!btp]
    \begin{subfigure}[t]{0.48\textwidth}
        \includegraphics[width=3in]{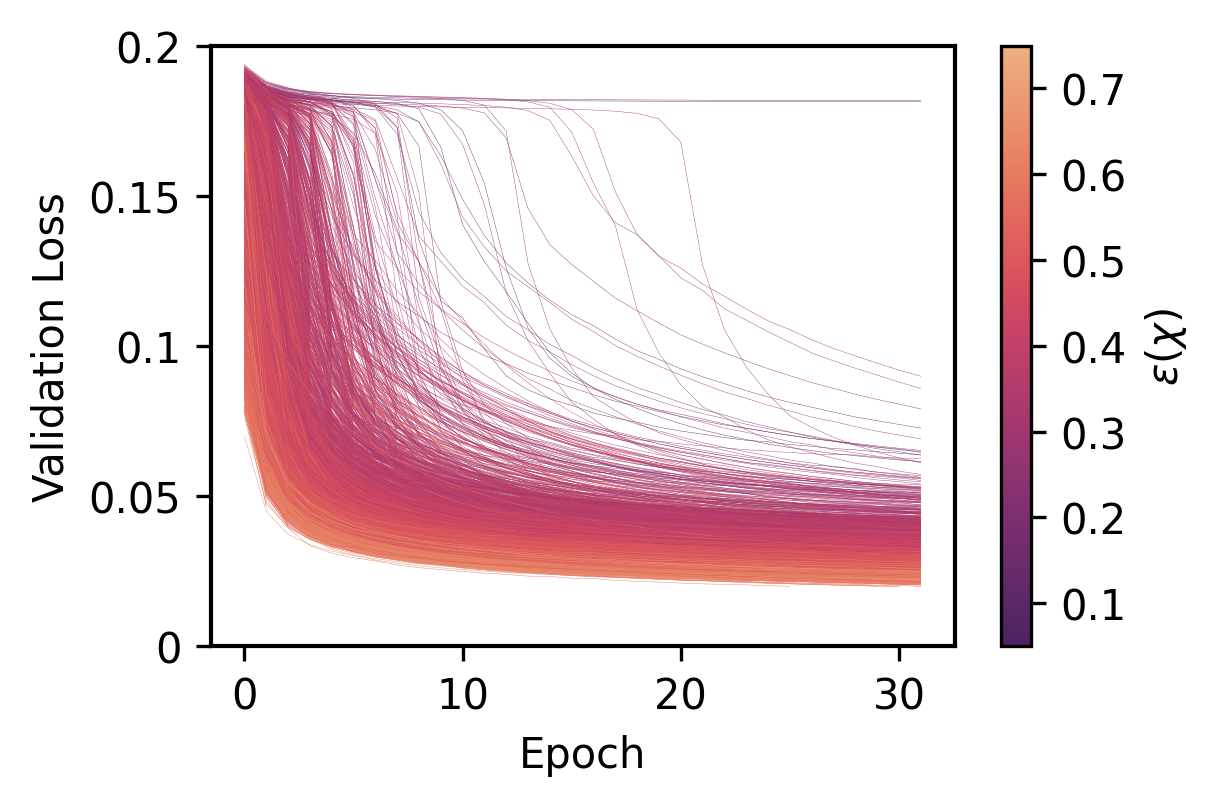}
    \end{subfigure}
    \begin{subfigure}[t]{0.48\textwidth}
        \includegraphics[width=3in]{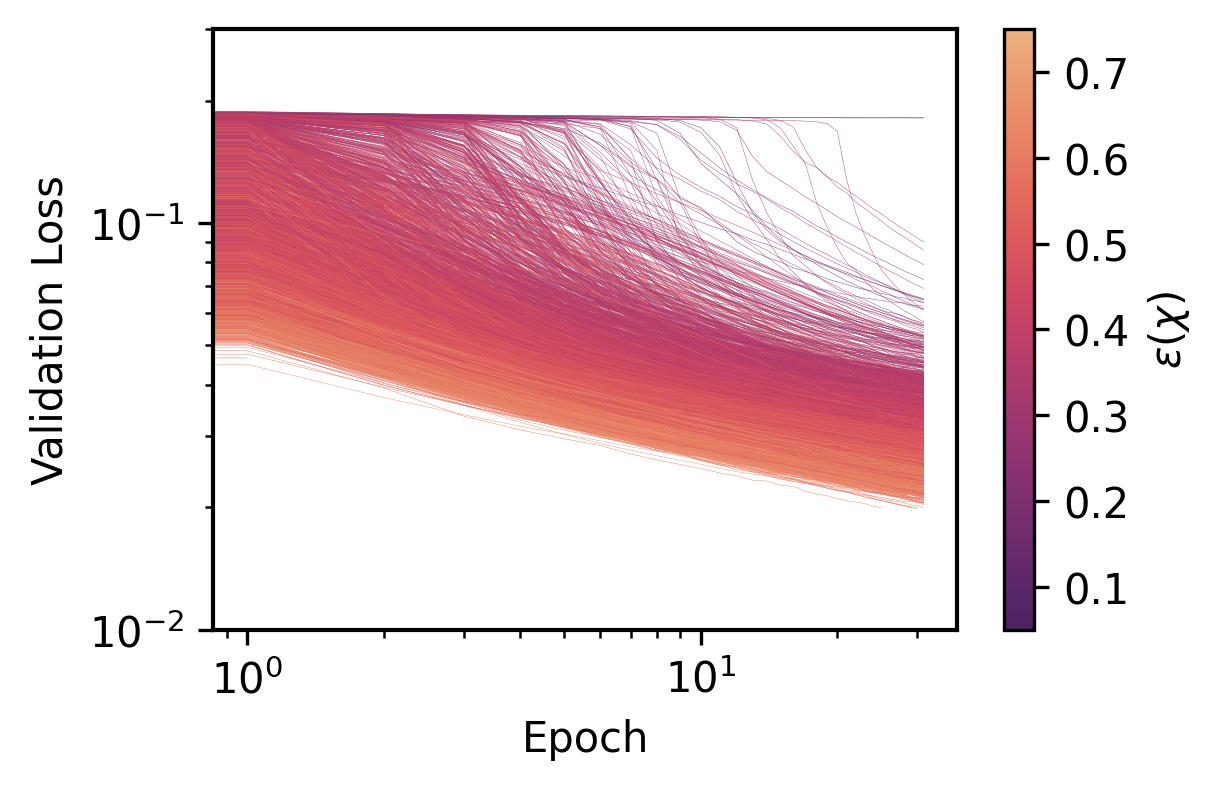}
    \end{subfigure}
    \caption{Model training curves in linear and logarithmic scales, respectively, for  neural networks trained and evaluated with on-the-fly aberration sampling (corresponding to Fig. 5 of main text). Training curves colored by $\epsilon(\chi)$ as measured on training dataset aberrations. }
    \label{sfig:eps_convergence_rate}
\end{figure}

\newpage
\section{Neural network spectral response}

\subsection{Dataset generation and model training}

\begin{table}[bhpt]
    \centering
    \begin{tabular}{r r r}
    \hline
     Structure & Space Group & Lattice parameters (\AA) \\
    \hline
     ZnS  & 186 & 3.8227 (a), 6.2607 (c) \\
     CdSe & 186 & 4.2985 (a), 7.0152 (c) \\
     Cu  & 225 & 3.6157\\
     Ag  & 225 & 4.0782\\
     C & 227 & 3.5673 \\
     Si & 227 & 5.430 \\
     Fe3O4 & 227 & 8.394\\
     Co3O4 & 227 & 8.065 \\
     Fe & 229 & 2.866 \\
     \hline
    \end{tabular}
    \caption{Lattice structures used for nanoparticle generation in structure varying datasets.}
    \label{tab:structures}
\end{table}

To measure the OOD response with respect to atomic structure domain shift, we generate a final set of datasets corresponding to a series of image datasets which primarily vary in structure instead of imaging conditions. We expand our nanoparticle structures to vary between eight different crystalline lattices and chemistries while keeping the nanoparticle morphologies and orientations the same as our previously sampled CdSe nanoparticles; no crystalline defects are sampled for these structures. In the same evaluation cohort we generate a secondary structure varying images by replacing the amorphous substrate with amorphous Si$_3$N$_4$, which is generated analogously to the amorphous carbon using a minimum Si-Si, Si-N, and N-N bond lengths of 3.01\AA, 1.73\AA, and 2.83\AA, respectively, and a reduced density of 2.0g/cm$^3$ \cite{UMESAKI1992120}.  Lastly, we also generate a third set of image datasets using the same nanoparticle lattices with controlled chemistry by setting all monatomic unit cells to Zn and all two-atom unit cells to ZnS chemistries, which we use to attempt to isolate the effect of structural variation from chemical variation.  For the imaging conditions, we generate images with a fixed spherical aberration of $C_{40} = 40 \mu\text{m}$ and defocus chosen by setting passband orders to $n=k/4$ for $k=5,6,9,13$; no other aberrations were applied. Electron dose and focal spread are again set to 300e-/\AA$^2$ and 10\AA, respectively.

An additional 4608 networks were trained to measure the effect of changes in structural distribution on model performance, 2048 each upon the datasets with the original nanoparticle lattices on amorphous carbon and Si$_3$N$_4$, and 512 models on the monochemistry dataset. To accelerate convergence and study performance at training convergence, we utilized the Adam optimizer and trained models with a learning rate of 1e-4, 1e-4, and 5e-5, respectively, a batch size of 16, and a learning rate decay factor of 0.99; model training was cut off after model after surpassing a loss of 0.05 above the optimum or 5 max epochs. 

\begin{figure}[t]
    \begin{subfigure}[t]{0.48\textwidth}
        \includegraphics[width=3in]{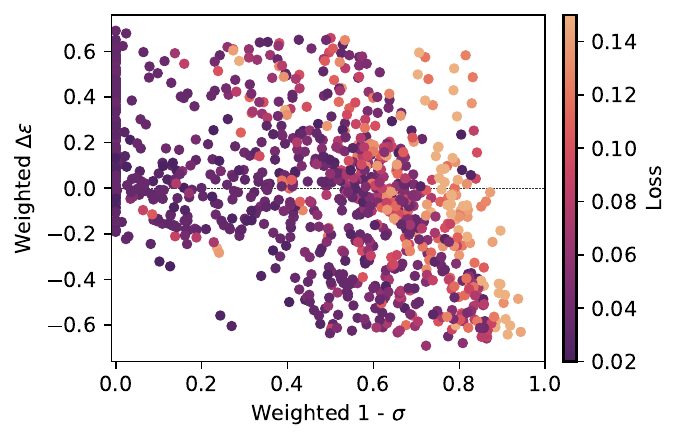}
    \end{subfigure}
    \begin{subfigure}[t]{0.48\textwidth}
        \includegraphics[width=3in]{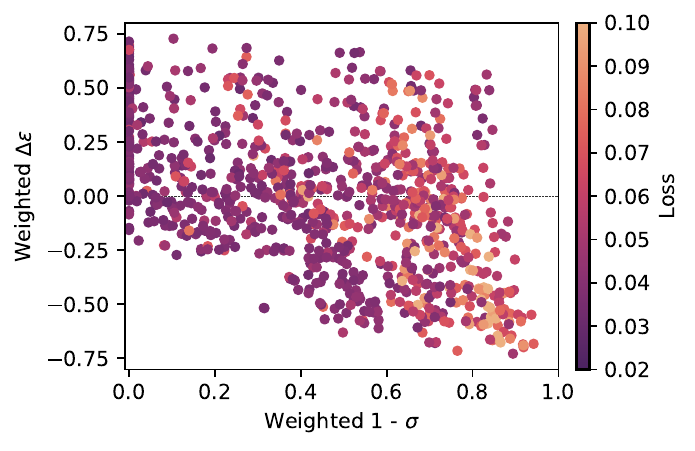}
    \end{subfigure}
    \caption{Out-of-distribution loss with respect both to nanoparticle structures and imaging conditions, utilizing normalized lattice structure factor-weighted $\epsilon$ and $\sigma$ metrics for (left) original nanoparticle structures and (right) nanoparticles with controlled chemistry.}
    \label{sfig:weighted_es}
\end{figure}

Spectral weights proportional to approximate structure factors for each nanoparticle distribution can be measured by averaging the Fourier transform of output wave intensities from multislice simulations. These approximate structure factors can be normalized and used as weighting measures $\omega$ for the $\epsilon$ and $\sigma$ metrics, i.e., $T(\mathbf{q})$ becomes $\tilde{T}(\mathbf{q}, \omega) = \omega(\mathbf{q}) T(\mathbf{q})$. When the weighted measures are employed, we can reorganize the (collapsed) $\epsilon-\sigma$ scatter plot, as in supplemental Fig. \ref{sfig:weighted_es}; we utilize spectral weights which are further radially averaged, for accelerating computation. Radial averaging inherently cannot capture differences in symmetry between two systems. We also tested using non-averaged spectral weights, essentially, instead calculating, e.g. for the overlap metric

\begin{equation*}
    \sigma(\chi, \chi') = \mathbb{E}_{i,j} \left[ \frac{\int \abs{\omega_i(\mathbf{q})T(\mathbf{q})} \abs{\omega_j(\mathbf{q})T'(\mathbf{q}) \mathrm{d}\mathbf{q}} }{ \int \abs{\omega_i(\mathbf{q}) T(\mathbf{q})}^2 \mathrm{d}\mathbf{q}} \right]
\end{equation*}

where the expectation is taken pairwise over a series of sampled simulations from each structure dataset. This is comparatively expensive to compute but can inherently capture symmetry-sensitive overlap, though, in our testing, this did not notably improve or make more meaningful the result obtained in Fig. \ref{sfig:weighted_es}. 

\section{Empirical KL divergence between imaging datasets}

Taking inspiration from \cite{ben-davidAnalysis2006}, the divergence between datasets can be empirically bound using an optimal classifier which distinguishes the two datasets. Intuitively, data generated via dissimilar generating distributions should be easier to classify than data that are extremely similar. We thus seek an approximate measure of distance between two data generating distributions that is hopefully cheap and quick to acquire using image classification models.

We begin by adapting our segmentation model architecture for classification. We remove the upsampling section of the network, and instead concatenate the outputs from each of the downsampling blocks. The output is then compressed to a lower dimensionality, first via an average pooling layer with a kernel size of 8 pixels, and then via a randomly sampled orthogonal projection (sampled via QR decomposition on a Gaussian normal matrix). We utilize this portion of the model to project the data to a low rank, and the low rank projections themselves are used during training, to reduce the training time and cost of the classifier training. In our experiments, we trained classification models from scratch and after pre-training the analogous segmentation model on the CdSe nanoparticle segmentation. We saw no distinct benefits when using pre-trained weights for the image dataset projection, and utilized randomly initialized projection models for the actual KL divergence measurements. 

For classification, we utilized a simple fully connected multilayer perceptron network, with layer sizes of 1024, 1024, and 256, ReLU activation, and softmax prediction. Classification optimization was for 256 training epochs against the cross-entropy loss using stochastic gradient descent with a learning rate of 6e-3, a batch size of 64, a weight decay of 1e-3, and a learning rate reduction factor of 0.99. After training, we extra low-dimensional (i.e, rank 256) learned features from the model by running the image projections through the MLP and taking the output of the penultimate layer. We then project the learned features onto a low-rank manifold using the UMAP algorithm \cite{McInnes2018}, here, using a rank of 8 dimensions. After this final manifold projection, we can estimate the KL divergence between image datasets using finite sets of samples of our data generating process \cite{PrezCruz2008KullbackLeiblerDE} \footnote{We use an open-source implementation for calculating the KL divergence located at the following link:\\ https://gist.github.com/atabakd/ed0f7581f8510c8587bc2f41a094b518}.

\begin{figure}[t]
    \centering
    \includegraphics[width=0.8\linewidth]{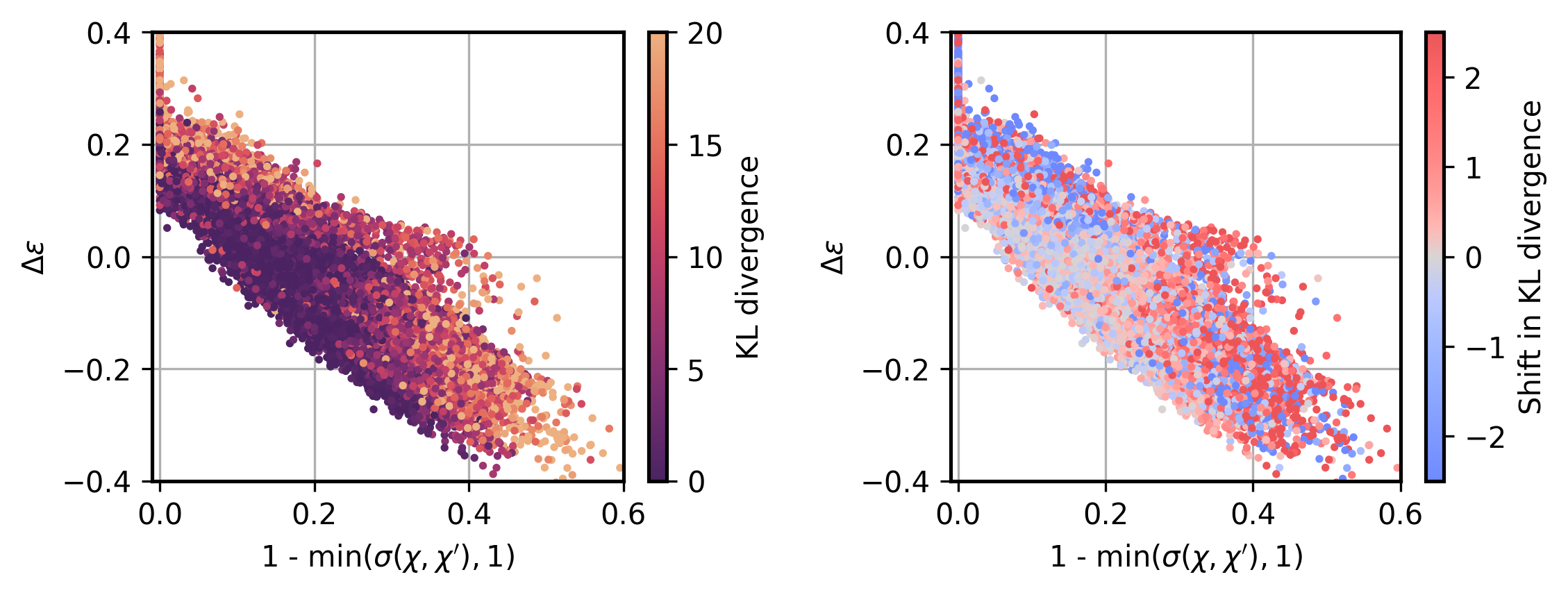}
    \caption{Dependence of KL divergence (left) and shifts in KL divergence (right) between training datasets and OOD data, relative to $\sigma(\chi, \chi')$ and $\epsilon(\chi)$. Aberrations of training and evaluation datasets were sampled up to 5th order (defocus, axial coma, two- and three-fold astigmatism, and spherical). Each marker corresponds to a unique training dataset--OOD dataset aberration pair.}
    \label{sfig:kl_es}
\end{figure}

KL divergence seems to spike both at small overlaps and negative $\Delta \epsilon$, and large overlaps and positive $\Delta \epsilon$, and seems informative that the data have changed significantly in their information content. However, it does not seem predictive as to the magnitude of the change in OOD loss, nor the direction of change. In testing, we found relative measurements of KL divergence to be fairly consistent and insensitive to the final classification performance, training hyperparameters, model ranks, UMAP manifold ranks and whether or not weights were reused from a model that was trained to perform segmentation or randomly initialized. We note that these measures do not provide a calibrated measure of distance; later, when discussing results comparing dataset distance to the induced loss out-of-distribution, it is important to consider that true calibration of such divergence or distance measures to (a bound of) the out-of-distribution induced loss would likely require a more precise data distribution divergence measure. Importantly, to get an approximate measure of dataset difference, the classifier we train does not even need to be necessarily good to obtain a measure of dataset difference--the data might be similar and thus hard to distinguish--but it could be important to train as good of a possible of a classifier as possible, in order to bound the dataset distance as tightly as possible. Nonetheless, we do not provide theoretical justification for why this approach seems to work, though, the results herein seem to be smooth and consistent for our data.

\section{Hessian property determination}

Measures of the model loss Hessian with respect to perturbation of the parameters are typically difficult to access and expensive to compute; however, some measures of hessian properties can be measured accurately with cheaper, stochastic methods, namely, stochastic Lanczos quadrature, for the Hessian eigenspectum density, and Hutchison's method, for trace estimation. For our second-order analysis, we perform measures of eigenspectrum density and estimate the full Hessian via a low-rank dominant subspace by calculating the top K eigenvectors via power iteration. We use PyHessian\cite{yaoPyHessianNeuralNetworks2020} to perform this analysis with some slight modifications to facilitate recording such metrics more fault-tolerant at scale. 

An additional 48 U-Net models were trained from scratch on the -15nm, 0nm, and +15nm CdSe defocus datasets in order to perform second order analysis of model loss functions. For these models, we save the model weights at each epochs, and optimize weights using stochastic gradient descent with a learning rate of 2.3e-3, a batch size of 16, and a learning rate decay of 0.98. Training conditions and OOD measurements are otherwise performed similarly as before. After model training, we measure the Hessian eigenspectrum, shown in supplemental Fig. \ref{sfig:hessian_esd}, and the top $K$ eigenvectors for each model after epochs 1, 4, 16, and 64 with respect to all eleven defocus datasets. Due to computation constraints and the size of the models, we use $K=12$ for the subspace estimation. As training continues, the Hessian eigenspectra tend to shift towards more positive values, and the overall distribution overlap for the spectra tends to be the best for datasets for which models generalize well to OOD.

\begin{figure}
    \centering
    \includegraphics[width=0.8\linewidth]{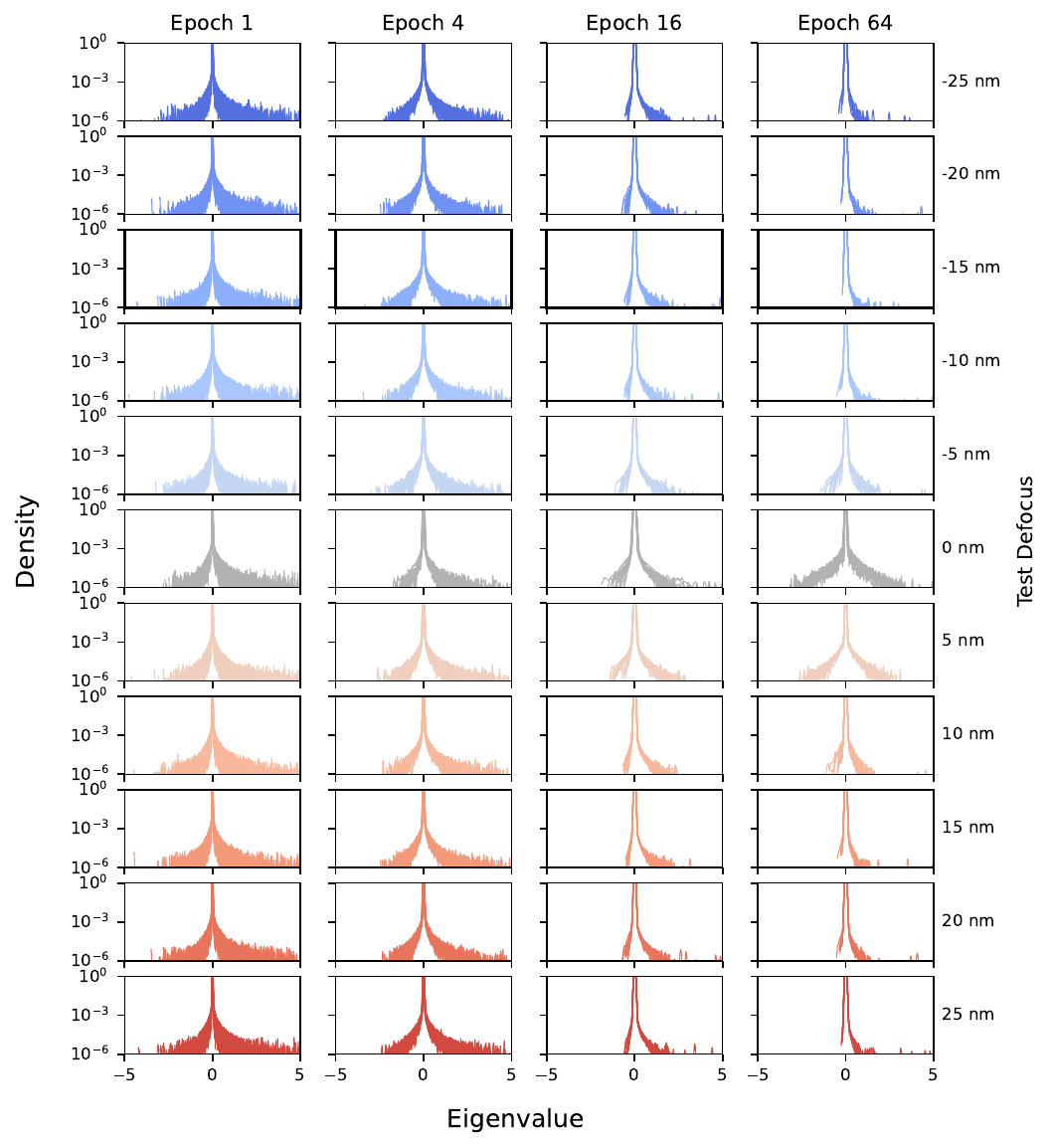}
    \caption{Evolution of Hessian eigenspectra throughout training for U-Net segmentation models trained on -15nm CdSe defocus data. }
    \label{sfig:hessian_esd}
\end{figure}

While the eigenspectra inform us about the overall curvature of the loss landscape, and give strong signals that loss landscapes evolve in degree of curvature in similar ways throughout training, the eigenspectra do not inform us about whether or not the comparative loss landscapes share dominant subspaces, that is, whether or not the same directions share curvature and thus whether or not the neural networks might share latent features across data domains. To compare the low rank Hessian subspaces, we use the Grassmanian metric, which is the squared sum of the principle angles between two subspaces. For example, in 2D, the principle angle between two subspaces--here, two lines passing through the origin--is simply the angle between the lines. In higher dimensions, the principal angles between two subspaces represented by the matrices $M \in \mathbb{R}^{m, k}$ and $M'\in \mathbb{R}^{m, k}$, whose columns are orthonormal basis vectors of each subspace, can be measured via the singular value decomposition

$$M{M'}^T = U \Sigma V^T$$

where the singular values $\sigma_i$ are the principal angles between the subspaces. Thus, the Grassmanian $Gr(M, M') = \sqrt{\sum_i \arccos(\sigma_i)^2}$ can be relatively cheaply computed. We measure the Grassmanian between low-rank eigenbases of our neural networks to get a sense of the relative overlap in manifold subspaces between neural network loss functions conditioned on different datasets. For improved numerical precision, we first re-orthonormalize the eigenvectors $M, M'$ via a modified Gram-Schmidt procedure, and then take the outer product between the two bases before finally calculating the SVD, which is fairly inexpensive due the fairly small size of $K$. We share the evolution of the Grassmanians between the loss Hessians on OOD datasets in supplemental Fig. \ref{sfig:hessian_gr}. As training continues, the loss subspaces tend to align across datasets for which the neural networks generalize well; the degree of alignment depends on the training dataset, and the 0nm (in focus) training data seems to induce the best subspace alignment.

\begin{figure}
    \centering
    \includegraphics[width=0.8\linewidth]{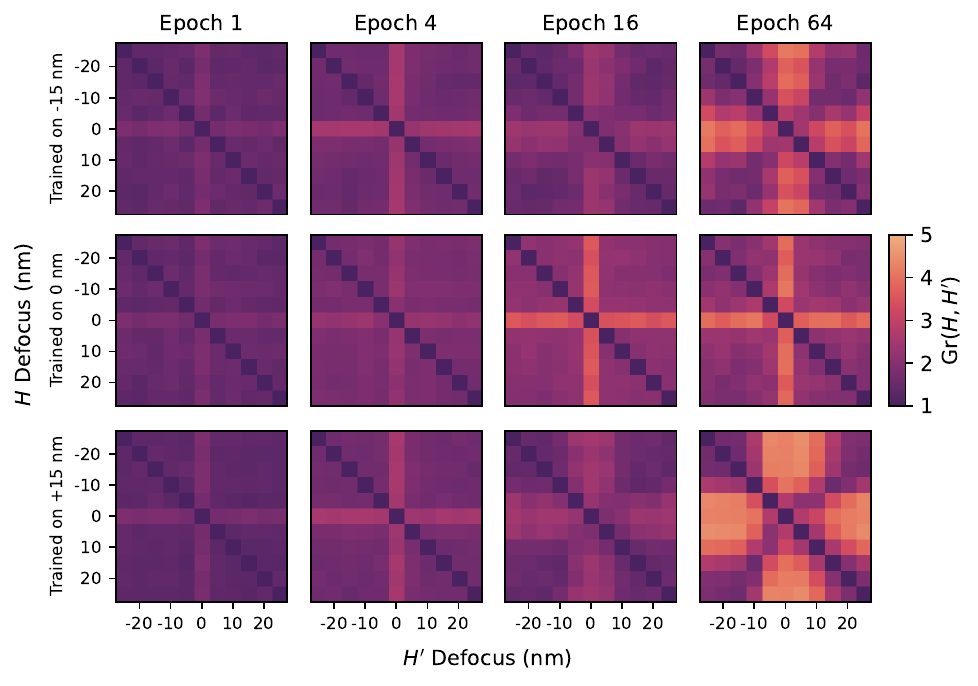}
    \caption{Evolution of Hessian Grassmanian throughout training for U-Net segmentation models trained on -15nm (top), 0nm (middle), and +15nm (bottom) defocus CdSe data. Mean value across 16 models for each data point.}
    \label{sfig:hessian_gr}
\end{figure}

We note here that we are essentially attempting to make measures of the local curvature of the loss landscape in parameter space. Further, this loss landscape is measured approximately via a sample of data--i.e., when considering these stochastic measures, we can consider such measures to be conditional on the distribution of the test data. Near local optima of the model, the loss landscape curvature dominates the behavior of the loss function, and thus, comparison of Hessian measures can provide some more quantitative insights into the extent of OOD performance degradation and possible alignment of learned representations and loss landscapes across different datasets. Leaning back into an information-theoretic and data-centric perspective, the loss Hessian is strongly related to the Fisher information matrix \cite{bottcherVisualizingHighdimensionalLoss2024}. Given bounded statistical divergences between data generating distributions, such as the KL divergence, we believe there could be potential analytical routes utilizing the Fisher information matrix for bounding differences between loss landscapes between datasets and thus OOD loss via measures of Hessian properties.

\bibliographystyle{ieeetr}
\bibliography{refs}